\documentclass[lettersize,journal]{IEEEtran}
\usepackage{amsmath,amsfonts}
\usepackage{algorithm}
\usepackage{algpseudocode}
\usepackage{array}
\usepackage[caption=false,font=normalsize,labelfont=sf,textfont=sf]{subfig}
\usepackage{textcomp}
\usepackage{stfloats}
\usepackage{url}
\usepackage{verbatim}
\usepackage{graphicx}
\usepackage{color}
\usepackage{cite}
\usepackage{booktabs}
\usepackage{array}
\usepackage[hidelinks]{hyperref}          
\usepackage[nameinlink,capitalize,noabbrev]{cleveref} 
\usepackage{pifont}
\newcommand{\cmark}{\ding{51}} 
\begin{document}

\title{CARGO: Carbon-Aware Gossip Orchestration in Smart Shipping}

\author{Alexandros S. Kalafatelis,~\IEEEmembership{Graduate Student Member,~IEEE,} Nikolaos Nomikos,~\IEEEmembership{Senior Member,~IEEE,} Vasileios Nikolakakis, Nikolaos Tsoulakos, and Panagiotis Trakadas
\thanks{A. S. Kalafatelis, N. Nomikos, V. Nikolakakis, and P. Trakadas are with the Department of Ports Management and Shipping, National and Kapodistrian University of Athens, 34400 Euboea, Greece, {\it Emails: {\tt \{alexkalafat,nomikosn,vnikolak,ptrakadas\}@uoa.gr}.}}
\thanks{N. Tsoulakos is with Laskaridis Shipping Co. Ltd, 14562, Kifisia Greece, {\it Email: {\tt tsoulakos@laskaridis.com}.}}
\thanks{Manuscript received March 29, 2026.}}

\maketitle

\begin{abstract}
Smart shipping operations increasingly depend on collaborative AI, yet the underlying data are generated across vessels with uneven connectivity, limited backhaul, and clear commercial sensitivity. In such settings, server-coordinated federated learning (FL) remains a weak systems assumption, depending on a reachable aggregation point and repeated wide-area synchronization, both of which are difficult to guarantee in maritime networks. A serverless gossip-based approach therefore represents a more natural approach, but existing decentralized methods still treat communication mainly as an optimization bottleneck, rather than as a resource that must be managed jointly with carbon cost, reliability, and long-term participation balance. In this context, this paper presents CARGO, a carbon-aware gossip orchestration framework for smart-shipping deployment. CARGO separates learning into a control plane and a data plane. The data plane performs local optimization with compressed gossip exchange, while the control plane decides, at each round, which vessels should participate, which communication edges should be activated, how aggressively updates should be compressed, and when recovery actions should be triggered. These decisions are guided by compact telemetry that combines predictive utility, carbon-related cost proxies, and participation-history signals. We evaluate CARGO under a predictive-maintenance scenario using operational bulk-carrier engine data and a trace-driven maritime communication protocol that captures client dropout, partial participation, packet loss, and multiple connectivity regimes, derived from mobility-aware vessel interactions. Across the tested stress settings, CARGO consistently remains in the high-accuracy regime while reducing carbon footprint and communication overheads, compared to accuracy-competitive decentralized baselines. Overall, the conducted performance evaluation demonstrates that CARGO is a feasible and practical solution for reliable and resource-conscious maritime AI deployment.

\end{abstract}


\section{Introduction}
\IEEEPARstart{M}{aritime} transport underpins global trade, but its digitalization and decarbonization agendas are now increasingly intertwined. Shipping carries over 80\% of world trade by volume, while the International Maritime Organization (IMO) has set progressively stricter decarbonization ambitions, including at least a 40\% reduction in carbon intensity by 2030 and net-zero greenhouse gas (GHG) emissions from international shipping by 2050~\cite{united2024review, mepc20232023}. In parallel, predictive maintenance (PdM) is becoming a critical capability for fleet operations, since failures in propulsion and auxiliary systems directly affect safety, schedule reliability, fuel consumption, and operating cost~\cite{ellefsen2019comprehensive, LIANG2024117619}. For maritime PdM, therefore, it is no longer sufficient to ask whether a learning system is accurate; one must also ask whether it can be deployed efficiently and robustly across vessels operating under diverse connectivity and resource conditions~\cite{jiang2025joint, kalafatelis2025fluid, pei2024review}.

These deployment constraints are particularly acute at fleet scale. Vessel data are naturally distributed, wide-area communication is heterogeneous and often intermittent, and reliance on a shore-side coordinator is not always possible in real shipping operations. This makes decentralized, serverless training especially attractive. Classical server-coordinated federated learning (FL) has been highly successful in terrestrial edge settings, yet its star topology depends on a reliably connected aggregation point. In contrast, gossip-style decentralized learning can exploit the peer-to-peer structure of opportunistic maritime contacts and continue training without requiring every round to terminate at a central server~\cite{mcmahan2017communication,alqurashi2022maritime,wei2021hybrid,hegedHus2019gossip}. For smart-shipping deployment, this is a system-level design choice rather than a minor implementation detail.

However, existing decentralized methods are still insufficient for this setting. Canonical methods, such as decentralized parallel stochastic gradient descent (D-PSGD) and stochastic gradient push (SGP) focus on consensus and convergence over a communication graph, but they do not explicitly decide which vessels should participate, which links should be used, or when communication should be curtailed or reinforced under operational conditions~\cite{lian2017can,assran2019stochastic}. Communication-efficient variants reduce payload size through compression, and sparse-gossip methods reduce the number of exchanged peers~\cite{koloskova2019decentralized, tang2022gossipfl}. Yet these methods still treat communication primarily as a bandwidth or optimization issue. As a consequence, they do not jointly account for predictive utility, time-varying carbon conditions, long-term participation balance, and packet losses, leveraging robust control mechanisms. For maritime PdM, that omission matters, as a method that reaches strong accuracy only through dense synchronization, repeated use of the same favorable vessels, or communication under carbon-intensive conditions is difficult to employ under real operational conditions.

Considering current shortcomings in collaborative AI-based smart shipping, this paper presents {CARGO}, a {c}arbon-{a}wa{r}e {g}ossip {o}rchestration framework for decentralized smart-shipping deployment. CARGO separates learning into a control plane and a decentralized data plane. The data plane retains standard local optimization with compressed gossip exchange, while the control plane decides at each round which vessels should participate, which communication edges should be activated, how aggressively updates should be compressed, and when reliability actions, such as resynchronization should be triggered. These decisions are guided by a compact telemetry interface that combines utility-related signals, carbon- and energy-related signals, and participation history signals. In this way, CARGO turns decentralized learning from a fixed communication routine into an online orchestration problem tailored to the decentralized nature of maritime operation. The main contributions of this paper are as follows:
\begin{enumerate}
    \item We formulate decentralized maritime operations under intermittent connectivity, as an online orchestration problem over a time-varying communication graph, coupling predictive utility, carbon-aware resource use, and participation-aware control.
    \item Then, CARGO, a control-plane/data-plane architecture is introduced, jointly scheduling node participation, edge activation, communication compression, and recovery actions for compressed gossip learning under maritime operating constraints.
    \item A trace-driven validation protocol for smart-shipping learning is developed, capturing client-availability dropout, partial participation, packet loss, and multiple connectivity regimes derived from mobility-aware vessel interaction patterns with port context.
    \item An empirical evaluation on operational bulk-carrier engine data under trace-driven maritime communication stress is presented, showing that CARGO maintains competitive predictive quality while reducing carbon and communication overheads over accuracy-driven decentralized baselines.
\end{enumerate}

The rest of the paper is organized as follows. Section~II reviews related work and positions CARGO with respect to decentralized learning, communication-efficient optimization, carbon-aware systems, and maritime AI. Section~III presents the system model and formal problem formulation. Section~IV describes the CARGO orchestration framework and its round-wise control policy. Section~V details the experimental setup, including the maritime validation protocol and baseline methods. Section~VI reports the empirical results, and Section~VII concludes the paper.

\section{Related Work}
\label{sec:related_work}
This section discusses the current state-of-the-art in areas relevant to our study, identifying current shortcomings that we aim to address through the proposed CARGO framework. We organize the discussion around five themes, i.e., decentralized and gossip-based learning, communication-efficient and reliability-aware distributed learning, carbon- and energy-aware ML systems, AI for smart ships and maritime predictive maintenance, and the resulting gap that motivates CARGO. The central point is that CARGO is not introduced as another generic optimizer variant but as a smart-shipping orchestration method, where communication intermittency, carbon-aware operation, and participation control must be handled jointly.

\vspace{-4mm}
\subsection{Decentralized and Gossip-Based Federated Learning}
\label{subsec:rw_decentralized}

FL has been a popular decentralized learning approach due to its privacy-preserving characteristics. In this context, server-coordinated FL is commonly traced to FedAvg, where clients perform local training and a central server periodically aggregates model updates~\cite{mcmahan2017communication}. This architecture has been highly successful in terrestrial edge settings, but its direct transfer to maritime operations is not straightforward. In fleet-scale shipping, communication conditions vary, leading to often intermittent connectivity with shore-side or cloud-based coordinator nodes~\cite{alqurashi2022maritime,wei2021hybrid, kalafatelis2024survey}. For this reason, the more relevant methodological neighborhood for the present paper is serverless decentralized learning, and in particular gossip-based learning~\cite{hegedHus2019gossip, hegedHus2021decentralized}.

Classical decentralized optimization provides the main starting point for serverless decentralized learning. Here, D-PSGD replaces centralized aggregation with neighbor-to-neighbor model mixing over a communication graph and shows that decentralized training can remain competitive with centralized parallel SGD~\cite{lian2017can}. SGP extends this approach to directed or asymmetric communication through push-sum style mixing~\cite{assran2019stochastic}. CHOCO-type methods reduce payload cost through compressed communication and error-compensated gossip while preserving convergence guarantees~\cite{koloskova2019decentralized}. These methods are central baselines for CARGO because they define the standard decentralized learning substrate on top of which orchestration can act.

A second line studies {sparser peer-to-peer information exchange}. GossipFL reduces communication by combining decentralization with adaptive sparse communication~\cite{tang2022gossipfl}. More broadly, Heged\H{u}s \emph{et al.} explicitly frame gossip learning as a decentralized alternative to federated learning, showing that serverless peer-to-peer learning can be competitive with server-based FL in dynamic environments~\cite{hegedHus2019gossip}. A recent work by Tundo \emph{et al.} further strengthens this perspective in edge settings, where gossip learning is used to improve robustness, reduce communication overheads, and tolerate failures in geographically distributed environments~\cite{tundo2025decentralized}. These works support the architectural choice made in CARGO, i.e., for a maritime environment characterized by opportunistic contacts and unreliable wide-area connectivity, serverless peer exchange is not an implementation detail, but a system-level design decision.

These works have two important gaps. First, existing decentralized and gossip-based methods are primarily optimizer-centric, i.e., they specify how models should be mixed, but rarely treat node participation, link activation, and compression as a joint control problem. Second, they are generally unaware of carbon intensity as a scheduling signal. These are precisely the areas where CARGO departs from the current state-of-the-art.

\vspace{-4mm}
\subsection{Communication-Efficient and Reliability-Aware Distributed Learning}
\label{subsec:rw_comm_reliability}
A large body of work treats communication as the principal bottleneck in distributed learning. One direction reduces payload size through quantization and sparsification. Quantized SGD (QSGD) introduces quantized gradient exchange with provable communication savings~\cite{alistarh2017qsgd}. Sparsified SGD with memory shows that aggressive sparsification can remain effective when accompanied by residual accumulation~\cite{stich2018sparsified}. Next, error-feedback methods generalize this principle by showing that biased compressors can recover the behavior of full-precision SGD when dropped information is explicitly reinjected~\cite{karimireddy2019error}. In decentralized settings, CHOCO-type methods integrate these ideas into compressed gossip updates~\cite{koloskova2019decentralized}.

Another line of research addresses imperfect communication where time-varying graphs, packet loss, and partial message delivery affect decentralized convergence. Here, such effects can be analyzed at the topology/mixing level, while another approach would be to modify the optimizer more explicitly. Soft-DSGD is representative here, as it models unreliable decentralized communication, adapting the update to partially received messages and link reliability estimates~\cite{ye2022decentralized}. This is an important step towards practical deployment, but its focus remains on robust decentralized optimization rather than on a broader orchestration policy.

What remains underdeveloped in this domain is the joint treatment of the operational decisions that matter in a smart-shipping scenario, i.e., who participates, which feasible links are used, how aggressively updates are compressed, and when recovery actions should be triggered. In this area, CARGO goes above current solutions, usually optimizing one or two of these dimensions at a time, treating them instead as coupled control variables within a single round decision process.

\vspace{-4mm}
\subsection{Carbon- and Energy-Aware ML Systems}
\label{subsec:rw_carbon_energy}
The need to evaluate machine learning systems beyond predictive accuracy is now well-established. Green AI argues that computational cost should be treated as a first-class evaluation axis rather than as an afterthought~\cite{schwartz2020green}. Henderson \emph{et al.} similarly advocate systematic reporting of energy and carbon footprints so that learning systems can be compared not only by accuracy, but also by their environmental cost~\cite{henderson2020towards}. These works provide the broader methodological basis for treating sustainability as part of the learning problem.

Within distributed learning, recent works have moved from general advocacy to explicit footprint modeling. Savazzi \emph{et al.} analyze the energy and carbon footprints of centralized learning, server-based FL, and consensus-driven decentralized learning, and show that communication efficiency strongly affects whether distributed training is environmentally preferable~\cite{savazzi2022energy}. Guerra \emph{et al.} compare server-based FL, gossip-style FL, and blockchain-enabled FL in terms of accuracy, communication overheads, convergence time, and energy cost~\cite{guerra2023cost}. On the algorithmic side, energy-aware client selection and carbon-aware FL configuration have begun to appear in server-coordinated FL, including GREED and Green FL~\cite{albelaihi2022green,yousefpour2023green}, while on the gossip-side, optimized gossip learning (OGL) adapts local training effort, i.e., epochs and exchange behavior to reduce energy consumption~\cite{dinani2024context}. However, OGL remains energy-aware rather than carbon-aware, using an infrastructure-assisted orchestrator, and does not address participation fairness and unreliable communication. This distinction is important for positioning CARGO, since to the best of our knowledge, no prior gossip-learning works exist, combining carbon-aware scheduling, fairness control, and packet-loss-aware maritime deployment.

\vspace{-4mm}
\subsection{AI for Smart Ships and Maritime Predictive Maintenance}
\label{subsec:rw_maritime}

The maritime domain is not merely an application backdrop; it changes the systems assumptions. Maritime communications span coastal terrestrial links, satellite backhaul, and opportunistic ship-to-ship connectivity, and are therefore far more heterogeneous than the connectivity model commonly assumed in terrestrial edge learning~\cite{alqurashi2022maritime,wei2021hybrid}. This makes always-on centralized coordination a comparatively weak systems assumption for fleet-scale analytics.

On the application side, recent maritime FL studies have established that collaborative learning is relevant for smart-shipping use cases, such as fuel consumption modeling, just-in-time arrival, and predictive maintenance~\cite{zhang2021adaptive,wang2023federated}. At the same time, maritime PdM surveys emphasize that propulsion and auxiliary systems are high-value targets for AI-based monitoring, but also highlight the scarcity of openly available datasets and the need for architectures that can be deployed under real communication and systems constraints~\cite{kalafatelis2025towards, zhang2022marine}. This is a major area for CARGO, supporting PdM with a novel deployment method, i.e., a carbon-aware decentralized orchestration layer over a gossip-style learning substrate.

Importantly, current FL studies remain predominantly server-coordinated and application-focused. To the best of our knowledge, there are no prior works on gossip-based maritime PdM, jointly addressing decentralized gossip learning, carbon-aware communication control, reliability stress, and fairness-aware participation.
\vspace{-4mm}
\subsection{State-of-the-Art Summary and CARGO Positioning}
\label{subsec:rw_gap}
Stemming from the previous discussion of relevant works Table~\ref{tab:rw_positioning} summarizes representative methods along the design dimensions that are most relevant to this paper. This initial comparison highlights the dimensions that are supported by CARGO towards enabling a novel collaborative AI framework for smart-shipping. 

\begin{table*}[t]
\centering
\caption{Positioning of representative methods against the main design dimensions targeted by CARGO.}
\label{tab:rw_positioning}
\scriptsize
\setlength{\tabcolsep}{3.6pt}
\renewcommand{\arraystretch}{0.96}
\begin{tabular}{lccccccc}
\hline
\textbf{Method} &
\textbf{Central} &
\textbf{Peer-to-peer} &
\textbf{Payload} &
\textbf{Carbon / energy} &
\textbf{Unreliable-link} &
\textbf{Sparse peer} &
\textbf{Smart-ship} \\
&
\textbf{coordinator} &
\textbf{decentralization} &
\textbf{compression} &
\textbf{aware scheduling} &
\textbf{handling / recovery} &
\textbf{communication} &
\textbf{target} \\
\hline
FedAvg~\cite{mcmahan2017communication}                & \cmark & --     & --     & --          & --     & --     & -- \\
D-PSGD~\cite{lian2017can}                             & --     & \cmark & --     & --          & --     & --     & -- \\
SGP~\cite{assran2019stochastic}                       & --     & \cmark & --     & --          & --     & --     & -- \\
CHOCO-SGD~\cite{koloskova2019decentralized}           & --     & \cmark & \cmark & --          & --     & --     & -- \\
GossipFL~\cite{tang2022gossipfl}                      & --     & \cmark & \cmark & --         & --     & \cmark & -- \\
Soft-DSGD~\cite{ye2022decentralized}                  & --     & \cmark & --     & --         & \cmark & --     & -- \\
q-FFL / q-FedAvg~\cite{li2019fair}                    & \cmark & --     & --     & --      & --     & --     & -- \\
GREED~\cite{albelaihi2022green}                       & \cmark & --     & --     & \cmark     & --     & --     & -- \\
GreenFL~\cite{yousefpour2023green}                   & \cmark & --  & --  & \cmark & --      & --  & -- \\
OGL~\cite{dinani2024context}                              & --  & \cmark & --  & \cmark & --      & --  & -- \\
GL for Edge Forecasting~\cite{tundo2025decentralized}            & --     & \cmark & --  & --        & \cmark & --  & -- \\
FL for Green Shipping~\cite{wang2023federated}        & \cmark & --     & --     & --  & --       & --     & \cmark \\
\textbf{CARGO (ours)}                                 & \textbf{--} & \textbf{\cmark} & \textbf{\cmark} & \textbf{\cmark} & \textbf{\cmark} & \textbf{\cmark} & \textbf{\cmark} \\
\hline
\end{tabular}
\end{table*}

Table~\ref{tab:rw_positioning} leads to two conclusions. First, the characteristics that CARGO leverages have been separately studied in previous works, i.e., decentralized mixing, compression, sparse peer exchange, fairness-aware learning, reliability-aware communication, and energy/carbon-aware scheduling. Second, none of the representative methods above jointly addresses them within one unified control framework, thus highlighting the gap that we aim to fill via CARGO.

\section{System Model and Problem Formulation}
\label{sec:system_model}
\subsection{Deployment Setting and System Assumptions}
\label{subsec:deployment_setting}
We consider a smart-shipping decentralized learning system with $N$ vessels, indexed by $\mathcal{V}=\{1,\ldots,N\}$. Each vessel $i\in\mathcal{V}$ stores a local dataset $\mathcal{D}_i$ and maintains a parameter vector $x_i(t)\in\mathbb{R}^P$ at communication round $t$. Communication is constrained by a time-varying graph
\begin{equation}
\mathcal{G}_t=\bigl(\mathcal{V},E_t^{\mathrm{cand}}),
\label{eq:graph}
\end{equation}
where ${E}^{\mathrm{cand}}_t$ is induced by the trajectory-derived maritime communication topology. Training proceeds in synchronized rounds where at each round, a subset of vessels performs local optimization and exchanges model information over a subset of feasible communication links.

The system is subject to three operational constraints. First, connectivity is intermittent, so the feasible neighborhood of a vessel changes over time. Second, availability is stochastic, in the sense that a vessel in the current topology may still be unavailable because of local workload, energy constraints, or operational disruption. Third, communication is imperfect, a link that is feasible at decision time may still experience packet loss during the actual exchange, so the delivered communication graph can be a strict subset of the graph that was activated by the controller. Before each round, the controller, treated here as a logical scheduling module rather than as a parameter server, has access to a lightweight telemetry summary derived from the previous learning state and the current system context.
\vspace{-4mm}
\subsection{Control and Data Plane Separation}
\label{subsec:control_data_plane}
CARGO is organized as a control plane operating over a decentralized-learning data plane. The data plane follows the standard pattern of local stochastic optimization with neighbor mixing used in decentralized SGD and push-sum style distributed training~\cite{lian2017can, assran2019stochastic}. The control plane does not alter the predictor architecture or the local optimizer. Instead, it decides which vessels participate, which links are activated, how exchanged messages are compressed, and how mixing weights are assigned. This separation is a central architectural property of CARGO, preserving compatibility with established decentralized learning updates while exposing an explicit orchestration layer that accounts for carbon intensity, participation fairness, and communication conditions.
\vspace{-4mm}
\subsection{Round Interface and Telemetry} \label{subsec:notation} Here, we provide details on the necessary interface and telemetry, required at each round. Let $a_i(t)\in\{0,1\}$ denote the availability indicator of vessel $i$ at round $t$, and let \begin{equation} \Omega_t=\{i\in\mathcal{V}:a_i(t)=1\} \label{eq:available_set} 
\end{equation} 
be the available-node set. When a participation fraction $f\in(0,1]$ is enforced, the target number of active vessels is 
\begin{equation} K_t=\min\!\bigl(\lceil fN\rceil,\ |\Omega_t|\bigr) \label{eq:active_target} 
\end{equation} For each vessel $i$, the control plane observes the telemetry vector 
\begin{equation} z_i(t)=\bigl(\ell_i(t),\,\delta_i(t),\,s_i(t),\,\rho_i(t),\,\chi_i(t)\bigr), \label{eq:telemetry} 
\end{equation} where $\ell_i(t)$ is the latest local loss, $\delta_i(t)$ is a disagreement-related local-state signal, $s_i(t)$ is the inactivity streak, $\rho_i(t)$ is the participation rate over a sliding window, and $\chi_i(t)$ is the carbon intensity in gCO$_2$/kWh. The candidate neighbor set of node $i$ is denoted by $N_i(t)=\{j:(i,j)\in E_t^{\mathrm{cand}}\}$. At round $t$, the control plane emits the decision bundle 
\begin{equation} \mathcal{D}_t= \bigl( A_t,\, E_t,\, W_t,\, \mathcal{C}_t,\, r_t \bigr), \label{eq:decision_bundle} \end{equation} 
where $A_t\subseteq\Omega_t$ is the active-node set, $E_t\subseteq E_t^{\mathrm{cand}}$ is the activated edge set, $W_t=[w_{ij}(t)]$ is the mixing matrix, $\mathcal{C}_t$ is the compression policy, and $r_t\in\{0,1\}$ is the resynchronization flag. 
\vspace{-4mm} 
\subsection{Optimization Objective} \label{subsec:objective} CARGO balances three competing goals at each round, i.e., selecting informative vessels for learning, limiting the carbon cost of computation and communication, and preventing systematic exclusion of particular vessels. We write the controller objective in the following generic form 
\begin{equation} \max_{\mathcal{D}_t}\; \mathcal{U}_t(\mathcal{D}_t) -\lambda_{\mathrm C}(t)\,\mathcal{C}_t^{\mathrm{cost}}(\mathcal{D}_t) -\lambda_{\mathrm F}(t)\,\mathcal{F}_t(\mathcal{D}_t), \label{eq:generic_objective} 
\end{equation} 
where $\mathcal{U}_t$ denotes aggregate learning utility, $\mathcal{C}_t^{\mathrm{cost}}$ denotes the carbon-related cost proxy, and $\mathcal{F}_t$ denotes the fairness penalty. The nonnegative multipliers $\lambda_{\mathrm C}(t)$ and $\lambda_{\mathrm F}(t)$ regulate the trade-off between predictive progress, sustainability, and participation balance. The feasible set is defined by the operational constraints of the current round, i.e., only available vessels may be activated, only feasible topology edges may be selected, the communication degree budget must be respected, and the mixing matrix must remain stochastic on the activated communication graph. Packet loss does not alter the control decision itself, but affects the delivered graph used by the data plane. 
\vspace{-4mm} 
\subsection{Problem Statement} \label{subsec:problem_statement} The overall problem is to design a sequential control policy $\Pi$ that maps the current topology, telemetry, and model state to a feasible round decision
\begin{equation} \Pi:\bigl(\mathcal{G}_t,\{z_i(t)\}_{i=1}^{N},\{x_i(t)\}_{i=1}^{N}\bigr)\mapsto \mathcal{D}_t, \label{eq:policy_map} 
\end{equation} so as to improve the final predictive performance while controlling cumulative carbon expenditure and maintaining participation fairness over the training horizon. Formally, we seek a policy that minimizes terminal validation loss, subject to budget and feasibility constraints
\begin{align} \min_{\Pi}\quad & \mathbb{E}\!\left[\mathcal{L}\!\left(\bar{x}(T)\right)\right] \label{eq:global_obj} \\ \text{s.t.}\quad & \sum_{t=1}^{T} C_{\mathrm{tot}}(t) \le B_{\mathrm C}, \label{eq:global_budget} \\ & \mathcal{D}_t \text{ is feasible for all } t=1,\ldots,T \label{eq:global_feasible} 
\end{align} Here, $\bar{x}(T)$ denotes the final consensus (or average) model, $\mathcal{L}(\cdot)$ is the validation loss, and $B_{\mathrm C}$ is a cumulative carbon budget. Because this problem is sequential, graph-coupled, and combinatorial, CARGO solves it approximately through a round-wise primal-dual control policy.

\section{CARGO Orchestration Framework}
\label{sec:cargo_framework}

\begin{figure*}[!t]
\centering
\includegraphics[width=\textwidth]{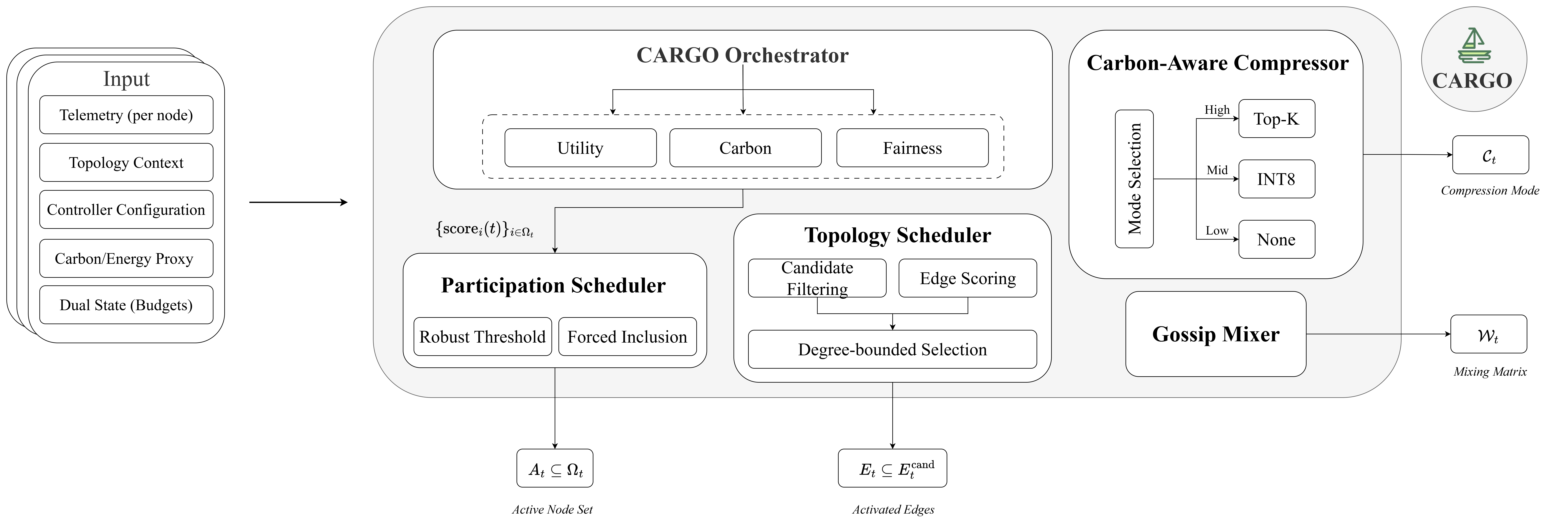}
\caption{CARGO control plane. The \emph{CARGO Orchestrator} combines utility, carbon, and fairness signals to compute per-node scores $\{g_i(t)\}_{i\in\Omega_t}$ over the available set. These scores drive the \emph{Participation Scheduler}, while the \emph{Topology Scheduler}, \emph{Carbon-Aware Compressor}, and \emph{Gossip Mixer} determine the activated graph $E_t$, compression policy $\mathcal{C}_t$, and mixing matrix $W_t$, respectively. The controller output at round $t$ is the decision bundle $\mathcal{D}_t=(A_t,E_t,W_t,\mathcal{C}_t,r_t)$, where $r_t$ denotes the resynchronization flag.}\label{fig:CARGO}
\end{figure*}

\subsection{CARGO Overview}
\label{subsec:cargo_overview}
Below, we provide the overview and control logic of the CARGO framework, clarifying how utility, carbon, and fairness signals are integrated to orchestrate node participation, topology formation, and communication efficiency. Fig.~\ref{fig:CARGO} summarizes the CARGO control plane. At the beginning of each round, the CARGO Orchestrator receives the current topology context, per-node telemetry, the carbon/energy proxy parameters, and the current dual state. It combines the utility, carbon, and fairness components to produce the per-available-node score set $\{g_i(t)\}_{i\in\Omega_t}$. These scores are consumed by the Participation Scheduler to determine the active-node set $A_t$. In parallel, the Topology Scheduler filters candidate links and selects a degree-bounded activated graph $E_t$, the Carbon-Aware Compressor assigns the communication mode for active senders, and the Gossip Mixer constructs the corresponding mixing matrix $W_t$. The resulting controller output is the round decision bundle $\mathcal{D}_t=(A_t,E_t,W_t,\mathcal{C}_t,r_t)$, which is then consumed by the decentralized data plane.
\vspace{-4mm}
\subsection{Participation Scheduler}
\label{subsec:participation_scheduler}

This subsection introduces the participation scheduling mechanism, determining the set of active nodes at each round by balancing utility, carbon cost, and fairness through adaptive scoring and thresholds. The Participation Scheduler operates on the per-node scores produced by the CARGO Orchestrator. For vessel $i$ at round $t$, the utility term is
\begin{equation}
U_i(t)=\frac{1}{1+\ell_i(t)}+\delta_i(t),
\label{eq:utility}
\end{equation}
which increases when the vessel remains informative for optimization or consensus.

The scheduler estimates the per-round carbon burden of activating node $i$ through a compute-side term and a communication-side term. Let $F(t)$ denote the estimated workload in FLOPs, $\tau_i$ the device throughput, and $P_i^{\mathrm{act}}$ the active power. The compute-side proxy is
\begin{equation}
C_{i,\mathrm{cmp}}(t)=
\left(\frac{F(t)}{\tau_i}P_i^{\mathrm{act}}\right)
\frac{\chi_i(t)}{3.6\times10^6}
\label{eq:compute_proxy}
\end{equation}
For communication, the transmitted-byte proxy is given inline by $B_{i,\mathrm{tx}}(t)=B_\theta\,\varrho_i(t)\,d_{\max}$, where $B_\theta$ is the dense model size, $\varrho_i(t)$ is the compression ratio, and $d_{\max}$ is the communication fanout cap. This induces a communication-side carbon term $C_{i,\mathrm{com}}(t)$, and the total proxy is
\begin{equation}
\Gamma_i(t)=C_{i,\mathrm{cmp}}(t)+C_{i,\mathrm{com}}(t)
\label{eq:total_proxy}
\end{equation}

To avoid chronic exclusion, the scheduler introduces a fairness penalty
\begin{equation}
\Phi_i(t)=
\bigl[s_i(t)-S_{\max}+1\bigr]_+
+
\bigl[\rho^\star-\rho_i(t)\bigr]_+,
\label{eq:fairness_penalty}
\end{equation}
where $S_{\max}$ is the maximum tolerated inactivity streak and $\rho^\star$ is the minimum target participation rate. The resulting score passed from the CARGO Orchestrator to the Participation Scheduler is
\begin{equation}
g_i(t)=
U_i(t)
-\lambda_{\mathrm C}(t)\Gamma_i(t)
-\lambda_{\mathrm F}(t)\Phi_i(t)
\label{eq:activation_score}
\end{equation}

The initial active set is formed through robust thresholding:
\begin{equation}
\vartheta_t=
\operatorname{median}\{g_j(t):j\in\Omega_t\}+\beta,
\label{eq:threshold}
\end{equation}
\begin{equation}
A_t^{(0)}=
\{i\in\Omega_t:g_i(t)>\vartheta_t\}
\label{eq:initial_active}
\end{equation}
Any vessel with inactivity streak at least $S_{\max}$ is then forced into the active set:
\begin{equation}
F_t=
\{i\in\Omega_t:s_i(t)\ge S_{\max}\},
\qquad
A_t\leftarrow A_t^{(0)}\cup F_t
\label{eq:forced_active}
\end{equation}
If the target cardinality in \eqref{eq:active_target} is not met, the set is filled or trimmed using a secondary efficiency ratio based on utility per unit carbon proxy.

\subsection{Topology Scheduler, Carbon-Aware Compressor, and Gossip Mixer}
\label{subsec:topology_compression_mixing}
\begin{figure}[!t] 
\centering 
\includegraphics[width=2.5in]{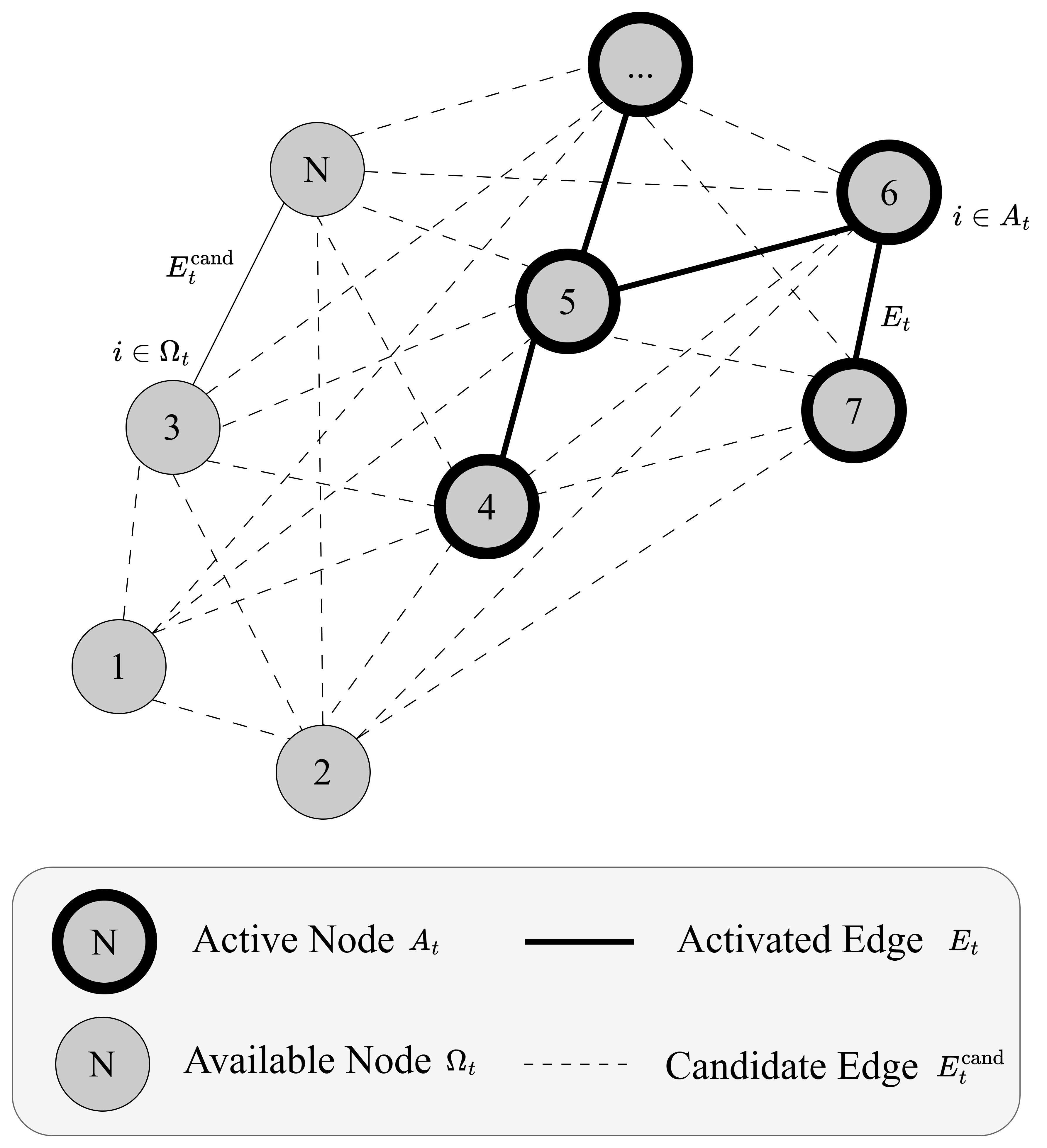} 
\caption{Round-level scheduling view. Starting from the available-node set $\Omega_t$ and the candidate edge set $E_t^{\mathrm{cand}}$ induced by the current topology snapshot, CARGO first selects the active-node set $A_t\subseteq\Omega_t$ and then activates a degree-bounded communication subgraph $E_t\subseteq E_t^{\mathrm{cand}}$.} \label{fig:Snap} 
\end{figure}

This subsection includes details on the coordinated design of topology selection, communication compression and decentralized aggregation, enabling efficient and carbon-aware information exchange across active vessels.
\paragraph{Topology Scheduler} For each active node $i\in A_t$, the candidate communication set is $S_i(t)=N_i(t)\cap A_t$. The Topology Scheduler ranks feasible edges according to an informativeness-to-cost ratio, 
\begin{equation} \psi_{ij}(t)= \frac{\left|\nu_i(t)-\nu_j(t)\right|} {\kappa_{ij}(t)+\varepsilon}, \label{eq:edge_priority} 
\end{equation} 
where $\nu_i(t)$ denotes the current node-level influence signal used for edge ranking and $\kappa_{ij}(t)$ denotes the communication-side carbon-cost proxy associated with activating edge $(i,j)$. Each active node keeps at most $d_{\max}$ neighbors with largest $\psi_{ij}(t)$, yielding a degree-bounded activated graph $E_t$. Fig.~\ref{fig:Snap} illustrates the distinction between the available-node set, the candidate topology edges, and the final activated subgraph used by the controller. 

\paragraph{Carbon-Aware Compressor} The Carbon-Aware Compressor selects the payload representation according to the sender’s carbon intensity. Let $\chi_{\mathrm L}$ and $\chi_{\mathrm H}$ denote the low- and high-intensity thresholds, then 
\begin{equation} \mathcal{C}_t(i)= \begin{cases} \text{None}, & \chi_i(t)<\chi_{\mathrm L},\\ \text{INT8}, & \chi_{\mathrm L}\le \chi_i(t)<\chi_{\mathrm H},\\ \text{Top-}K, & \chi_i(t)\ge \chi_{\mathrm H} \end{cases} \label{eq:compression_policy} 
\end{equation} 
where \textit{None} denotes no compression, i.e., dense transmission. 

\paragraph{Gossip Mixer} Given the activated graph $E_t$, the Gossip Mixer constructs a row-stochastic matrix using Metropolis weights, a standard local rule in distributed averaging and gossip-style consensus~\cite{xiao2004fast,boyd2006randomized}: 

\begin{equation} 
w_{ij}(t)= \begin{cases} \displaystyle \frac{1}{1+\max\{\deg_i(t),\deg_j(t)\}}, & (i,j)\in E_t,\ i\neq j, \\[8pt] \displaystyle 1-\sum_{j\neq i} w_{ij}(t), & i=j, \\[3pt] 0, & \text{otherwise}
\end{cases} \label{eq:metropolis} 
\end{equation}

\vspace{-4mm}
\subsection{Data-Plane Update With Compressed Gossip}
\label{subsec:data_plane_update}
Here, details are given regarding the data-plane update process, where compressed gossip-based exchanges are used to propagate model updates while mitigating communication overheads and ensuring convergence. Let $h_i(t)\in\mathbb{R}^P$ denote the local memory variable used for error compensation. For each active vessel $i\in A_t$, local training first produces an intermediate iterate $\widetilde{x}_i(t)$. The compressed-gossip step then proceeds as 
\begin{align} e_i(t) &= \widetilde{x}_i(t)-h_i(t), \label{eq:error_signal}\\ \widehat{e}_i(t) &= Q_i(t)\!\left(e_i(t)\right), 
\label{eq:compressed_error}\\ h_i(t+1) &= h_i(t)+\widehat{e}_i(t)
\label{eq:memory} 
\end{align} 
where $Q_i(t)$ is the compression operator induced by~\eqref{eq:compression_policy}. Packet loss determines the delivered-edge mixing matrix $\widetilde{W}_t=[\widetilde{w}_{ij}(t)]$, obtained from the activated matrix $W_t=[w_{ij}(t)]$ through 
\begin{equation} \widetilde{w}_{ij}(t)= \frac{w_{ij}(t)m_{ij}(t)} {\sum_{k=1}^{N} w_{ik}(t)m_{ik}(t)}, \label{eq:delivered_weights} 
\end{equation} whenever the denominator is nonzero. Otherwise, node $i$ retains full self-weight, i.e., $\widetilde{w}_{ii}(t)=1$ and $\widetilde{w}_{ij}(t)=0$ for $j\neq i$. Using $\widetilde{W}_t$, the mixed-memory state is 
\begin{equation} \bar{h}_i(t+1)=\sum_{j=1}^{N}\widetilde{w}_{ij}(t)h_j(t+1), \label{eq:mixed_memory} 
\end{equation} 
and the model update is 
\begin{equation} x_i(t+1)=\widetilde{x}_i(t)+\gamma\bigl(\bar{h}_i(t+1)-h_i(t+1)\bigr), \label{eq:state_update} 
\end{equation} 
where $\gamma$ is the gossip step size and $\widetilde{W}_t$ is the delivered-edge mixing matrix defined in~\eqref{eq:delivered_weights}. This update follows the error-feedback principle of compressed decentralized optimization~\cite{koloskova2019decentralized,karimireddy2019error}. Nodes not in $A_t$ keep both state and memory unchanged. \vspace{-4mm} 

\subsection{Reliability and Resynchronization} \label{subsec:reliability_resync} This part clarifies the impact of unreliable communications, introducing mechanisms to handle packet loss and maintain system stability through proper resynchronization. More specifically, packet loss acts on attempted transmissions and induces a delivered graph that is generally sparser than the activated graph selected by the controller. Let $m_{ij}(t)\in\{0,1\}$ denote the delivery indicator on link $(i,j)$. The delivered-edge mixing matrix $\widetilde{W}_t$ is defined in~\eqref{eq:delivered_weights} and is used directly in the compressed-gossip update. The empirical effective loss over a run is reported as \begin{equation} 
p_{\mathrm{eff}}= 1- \frac{\sum_t B_{\mathrm{del}}(t)} {\sum_t B_{\mathrm{att}}(t)}, 
\label{eq:effective_loss} 
\end{equation} 
where $B_{\mathrm{att}}(t)$ and $B_{\mathrm{del}}(t)$ denote attempted and delivered bytes, respectively. To reduce drift under communication impairment, CARGO exposes resynchronization through the flag $r_t$ in~\eqref{eq:decision_bundle}. In the reported experiments, resynchronization follows a fixed periodic rule, 
\begin{equation}
r_t=
\begin{cases}
1, & t \bmod R = 0,\\
0, & \text{otherwise}
\end{cases}
\label{eq:resync_trigger}
\end{equation}
where $R$ denotes the resynchronization interval in rounds. The value of $R$ is set by the runtime preset and remains fixed within each experiment family.
\vspace{-4mm}

\subsection{Dual Updates}
\label{subsec:dual_updates}
Finally, this subsection presents the update rules for the dual variables, dynamically regulating carbon budget and fairness constraints. The CARGO Orchestrator updates the adaptive multipliers $\lambda_{\mathrm C}(t)$ and $\lambda_{\mathrm F}(t)$ using projected subgradient steps. Let
\begin{equation}
\mathcal{C}_{1:t}=\sum_{\tau=1}^{t} C_{\mathrm{tot}}(\tau)
\label{eq:cumulative_carbon}
\end{equation}
denote cumulative carbon and $B_{\mathrm C}$ the nominal carbon budget. The normalized carbon gap is
\begin{equation}
g_{\mathrm C}(t)=
\frac{\mathcal{C}_{1:t}-B_{\mathrm C}}{B_{\mathrm C}+\varepsilon}
\label{eq:carbon_gap}
\end{equation}
Similarly, with average participation $\bar{\rho}(t)=\frac{1}{N}\sum_{i=1}^{N}\rho_i(t)$, the fairness gap is
\begin{equation}
g_{\mathrm F}(t)=
\bigl[\rho^\star-\bar{\rho}(t)\bigr]_+
\label{eq:fairness_gap}
\end{equation}
The dual updates are
\begin{align}
\lambda_{\mathrm C}(t+1) &=
\bigl[\lambda_{\mathrm C}(t)+\eta_{\mathrm C} g_{\mathrm C}(t)\bigr]_+,
\label{eq:dual_c}
\\
\lambda_{\mathrm F}(t+1) &=
\bigl[\lambda_{\mathrm F}(t)+\eta_{\mathrm F} g_{\mathrm F}(t)\bigr]_+
\label{eq:dual_f}
\end{align}

\algrenewcommand\algorithmicrequire{\textbf{Input:}}
\algrenewcommand\algorithmicensure{\textbf{Output:}}
\begin{algorithm}[t]
\caption{CARGO at round $t$}
\label{alg:cargo}
\scriptsize
\begin{algorithmic}[1]
\Require $\Omega_t$, $\mathcal{G}_t$, $\{z_i(t)\}_{i=1}^{N}$, $\{x_i(t),h_i(t)\}_{i=1}^{N}$, $\lambda_{\mathrm C}(t)$, $\lambda_{\mathrm F}(t)$
\Ensure $\mathcal{D}_t$, $\{x_i(t+1),h_i(t+1)\}_{i=1}^{N}$, $\lambda_{\mathrm C}(t+1)$, $\lambda_{\mathrm F}(t+1)$

\ForAll{$i\in\Omega_t$}
    \State Compute $U_i(t)$, $\Gamma_i(t)$, $\Phi_i(t)$, and $g_i(t)$ using \eqref{eq:utility}-\eqref{eq:activation_score}
\EndFor

\State Form $A_t$ using \eqref{eq:threshold}--\eqref{eq:forced_active} and cardinality correction

\ForAll{$i\in A_t$}
    \State Select up to $d_{\max}$ neighbors in $N_i(t)\cap A_t$ using \eqref{eq:edge_priority}
    \State Assign compression mode $\mathcal{C}_t(i)$ by \eqref{eq:compression_policy}
\EndFor

\State Construct $E_t$ and $W_t$ using \eqref{eq:metropolis}; set $r_t$

\ForAll{$i\in A_t$}
    \State Compute $\widetilde{x}_i(t)$ by local SGD
    \State Compute $e_i(t)$, $\widehat{e}_i(t)$, and $h_i(t+1)$ via \eqref{eq:error_signal}-\eqref{eq:memory}
\EndFor

\State Construct $\widetilde{W}_t$ from $W_t$ and the realized delivery mask using \eqref{eq:delivered_weights}

\ForAll{$i\in A_t$}
    \State Compute $\bar{h}_i(t+1)$ and $x_i(t+1)$ via \eqref{eq:mixed_memory}-\eqref{eq:state_update}
\EndFor

\ForAll{$i\notin A_t$}
    \State $x_i(t+1)\gets x_i(t)$ and $h_i(t+1)\gets h_i(t)$
\EndFor

\If{$r_t=1$}
    \State perform resynchronization over delivered links
\EndIf

\State Update $\lambda_{\mathrm C}(t+1)$ and $\lambda_{\mathrm F}(t+1)$ using \eqref{eq:dual_c}-\eqref{eq:dual_f}
\end{algorithmic}
\end{algorithm}

\section{Experimental Setup}
In this section, we describe the experimental setup and implementation details used to evaluate CARGO. More specifically, all experiments were executed on a local workstation featuring an Intel\textsuperscript{\textregistered} Core\texttrademark{} Ultra i9-275HX CPU (2.7\,GHz), 32\,GB RAM, a 64-bit operating system, and an NVIDIA GeForce RTX\,5070 GPU. The experimental pipeline was implemented in Python~3.12.3 using PyTorch~2.9.1 (CUDA~12.8). Training and evaluation were GPU-accelerated via CUDA (cuDNN~9.10.2). All decentralized experiments were run with identical software and measurement settings across all methods to ensure a consistent comparison.
\vspace{-4mm}
\subsection{Data, Preprocessing, Feature Engineering, and PdM Model}
\subsubsection{Dataset Description}
We evaluate CARGO using operational data from a bulk carrier operated by Laskaridis Shipping Ltd., collected from a two-stroke Main Engine (ME) rated at 12{,}009~HP and previously analyzed in~\cite{kalafatelis2025privacy}. The dataset contains 59{,}619 high-frequency time-series observations acquired from 759 shipboard sensor channels. The learning target is the cylinder-level exhaust gas temperature (EGT), a routinely monitored variable in marine PdM that reflects the engine’s thermal balance and combustion condition, being widely used for early detection of abnormal firing, injection faults, and heat-transfer degradation~\cite{cheliotis2020machine,ji2023deep,liu2022research, kalafatelis2025explainable}.

\subsubsection{Preprocessing, Feature Selection \& Engineering}
To prevent temporal leakage, we apply a contiguous 80/20 train/test split before any scaling, windowing, or feature ranking. Missing predictor values are imputed via forward fill followed by backward fill, while samples with missing target EGT are removed. Feature selection is performed on the training split only. In this context, we retain numeric predictors, exclude the target, and rank candidates by absolute Pearson correlation ($s_j=\left|\mathrm{corr}(x_j,y)\right|$), keeping the top-$9$ variables (Table \ref{tab:features_rationale}). Then, we augment the selected predictors with lightweight temporal structure, i.e., a one-step target lag $y_{t-1}$ and, for each selected $x_j$, the first difference $\Delta x_{j,t}=x_{j,t}-x_{j,t-1}$, as well as a 12-sample rolling mean $\bar{x}^{(12)}_{j,t}$. Also, a relative time index is included, while rows that are undefined by lagging/differencing are dropped. Predictors and target are standardized with separate StandardScaler transforms fitted on the training split and applied to the test split. Furthermore, supervised samples are formed with sliding windows of length $W$ and horizon $H$, based on $\mathbf{X}_t=\{\mathbf{x}_t,\ldots,\mathbf{x}_{t+W-1}\}, \quad 
y_t = y_{t+W+H-1}$, where unless stated otherwise, $W{=}30$ and $H{=}1$ was set, yielding a compact multivariate sequence per timestep consisting of the selected raw variables, their differenced and smoothed counterparts, the lagged target, and the relative time index.

\begin{table}[t]
\centering
\caption{Core engine variables used for ME cylinder EGT forecasting.}
\includegraphics[width=3in]{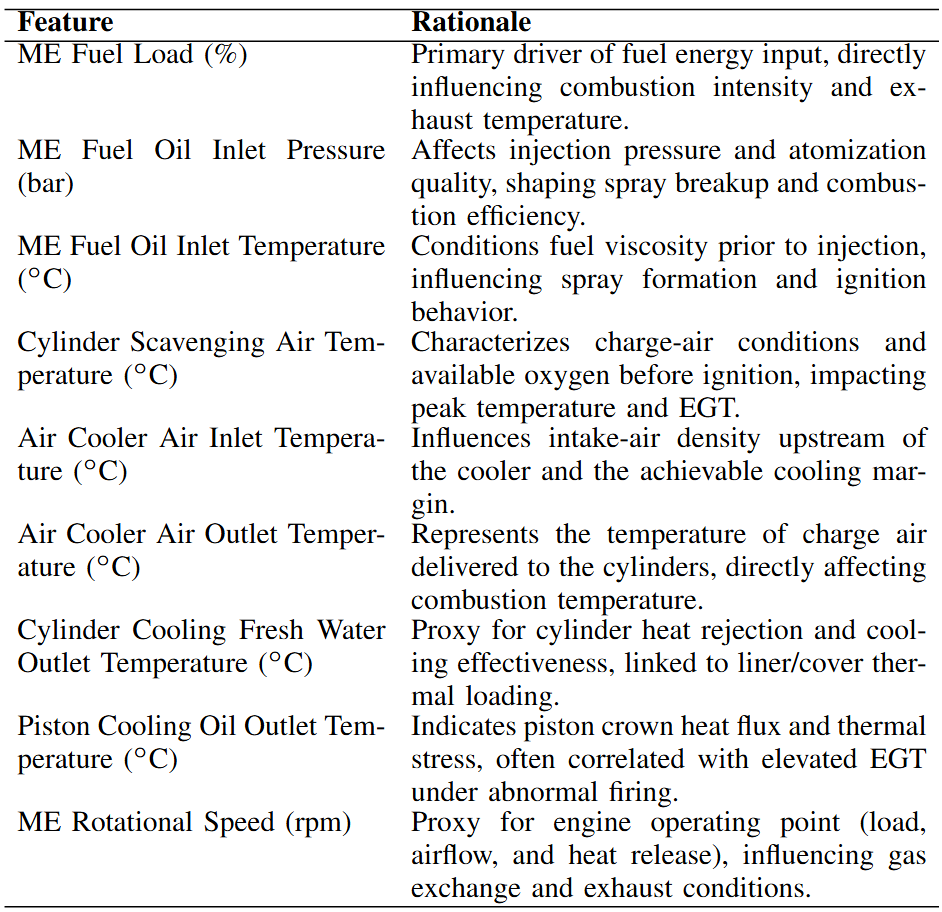}
\label{tab:features_rationale}
\end{table}

\subsubsection{PdM Model}
Towards ensuring a fair comparison, all decentralized experiments use the same PdM model, implemented as a sequence-to-one long short-term memory (LSTM) regressor. The backbone is a two-layer LSTM with dropout $0.1$ between recurrent layers and hidden size $h{=}128$, selected via a small sweep over $h\in\{64,128,256\}$. The output is converted to a single-step EGT forecast, using a compact two-layer multilayer perceptron (MLP) head, i.e., a linear projection that halves the hidden dimensionality ($h \rightarrow h/2$), a ReLU nonlinearity, and a final linear layer that maps to one scalar output ($h/2 \rightarrow 1$). The model is trained by minimizing mean squared error (MSE) between the predicted and ground-truth EGT values, computed in the standardized target space.
\vspace{-4mm}
\subsection{Decentralized Validation Protocol, Topology Regimes, and Baselines} 

\subsubsection{Decentralized Maritime Validation Protocol}
\begin{figure}[!t]
\centering
\includegraphics[width=3.55in]{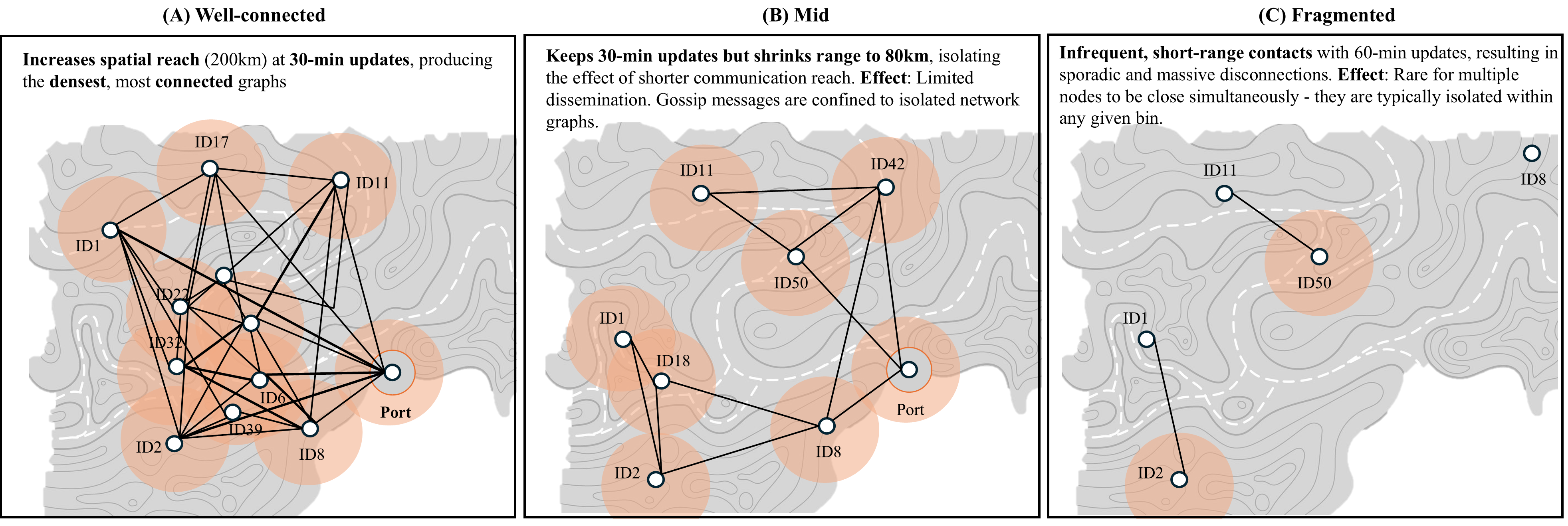}
    \caption{Visualization of maritime gossip topologies across three regimes: (A) Well-connected, (B) mid, and (C) fragmented.}
    \label{fig:topologies}
\end{figure}

We evaluate CARGO in a decentralized PdM setting where each vessel acts as a client holding local time-series data and exchanging model updates over an intermittently connected maritime network. Client-side instability is modeled through an availability dropout probability $p_d$, which removes a client from a given round, and a participation fraction $f$, which sub-samples from the available clients. Communication unreliability is modeled through an independent packet-loss probability $p$ applied to attempted transmissions. This protocol is intended to reflect the coverage gaps, link heterogeneity, and intermittent contact opportunities, being representative of maritime communications~\cite{alqurashi2022maritime,wei2021hybrid}. To avoid toy graph families while retaining experimental control, we construct AIS-style vessel trajectories with explicit port calls and derive time-varying contact graphs from vessel positions. Unless otherwise stated, we place $N{=}5$ vessels and $P{=}6$ ports within a 300\,km region, centered at $(37.0,-122.0)$ and simulate 48 base steps at 30-min resolution. Each vessel alternates between an in-port dwell phase (truncated Normal: mean 4 steps, std 1, minimum 1) and an underway phase towards a uniformly sampled destination port. Underway transit time is computed from great-circle distance and a service-speed draw centered around typical cruising values (mean 28\,km/h, std 6, truncated to $\ge$5\,km/h). To reflect reporting irregularity and missingness commonly observed in operational AIS feeds, we inject state-dependent gaps (in port versus at sea) and apply conservative gap-bridging during topology construction~\cite{emmens2021promises}. Connectivity is represented as a sequence of snapshot graphs with bin duration $\Delta t$. In snapshot $t$, an undirected edge $(i,j)$ is active if $d_{ij}(t)\le d_{\max}$, where $d_{ij}$ is the haversine distance. Vessels co-located at the same port are additionally connected as a dense local subgraph to represent near-shore/local infrastructure connectivity. Brief reporting gaps are handled by hold-last-position for up to one missing bin, without positional jitter. We evaluate three connectivity regimes: (i) well-connected $(30\,\mathrm{min},200\,\mathrm{km})$, (ii) mid $(30\,\mathrm{min},80\,\mathrm{km})$, and (iii) fragmented $(60\,\mathrm{min},80\,\mathrm{km})$. The transition well-connected$\rightarrow$mid isolates reduced spatial reach at fixed temporal resolution, while mid$\rightarrow$fragmented isolates reduced contact opportunities induced by coarser snapshotting at fixed range. Fig.~\ref{fig:topologies} visualizes the regimes, and Table~\ref{tab:regime_stats_n5} reports the corresponding connectivity statistics. These values are used to verify regime separation rather than hypothesis tests.


\begin{table}[t]
\centering
\caption{Connectivity statistics for the three topology regimes.}
\label{tab:regime_stats_n5}
\scriptsize
\setlength{\tabcolsep}{2.4pt}
\renewcommand{\arraystretch}{0.90}
\begin{tabular}{lccccc}
\hline
Regime & $\Delta t$ & $d_{\max}$ & Conn. & Med. & LCR \\
 & (min) & (km) & (cnt/rate) & comps & (mean) \\
\hline
Well-connected & 30 & 200 & 92/144 (0.639) & 1 & 0.883 \\
Mid            & 30 & 80  & 8/144 (0.056)  & 2 & 0.692 \\
Fragmented     & 60 & 80  & 5/72 (0.069)   & 2 & 0.686 \\
\hline
\end{tabular}
\end{table}

\subsubsection{Device Profile}
All clients are assigned the same low-power edge-device profile in the accounting model, so that differences in energy, carbon, and communication arise from the learning/control policy and network conditions rather than from heterogeneous hardware. In addition, we use a homogeneous edge profile representative of resource-constrained deployments often considered in recent edge studies, such as Jetson Nano-class devices, with active power $P_{\text{active}}{=}10$\,W, idle power $P_{\text{idle}}{=}1.5$\,W, compute throughput $\tau_{\text{dev}}{=}2.0\times10^{10}$ FLOPs/s, and uplink/downlink communication cost $\epsilon{=}2.0\times10^{-7}$ J/byte \cite{anuja2025end, cruz2023multi}. This choice is intentional, removing hardware heterogeneity as a confounder and allowing the evaluation to focus on the effect of CARGO's orchestration decisions under dropout, packet loss, and changing topology. 

\subsubsection{Client-Stress and Packet-Loss Protocol}
To assess robustness to client-side instability, we evaluate availability dropout $p_d \in \{0.2,0.5\}$ and participation fraction $f \in \{0.25,0.5,1.0\}$. Then, to evaluate robustness to unreliable communication, we consider packet-loss probability $p \in \{0,0.05,0.1,0.2\}$. Following prior network-impaired FL evaluations, lossy communication is modeled as independent per-message Bernoulli drops~\cite{cleland2022fedcomm}. Messages that are not received are excluded from mixing, and mixing weights are renormalized over delivered neighbors, so that each round still uses a valid stochastic combination. We report both attempted communication, defined as cumulative payload bytes before packet-loss masking, and delivered communication, defined as the subset of those bytes actually received after masking. This distinction is important because topology changes can alter effective convergence through mixing quality without necessarily inducing proportional changes in the attempted payload volume.

\subsubsection{Decentralized Baseline Schemes}
In terms of baseline schemes for comparisons purposes, CARGO is put against four decentralized alternatives, spanning the main design axes relevant to our setting. First, {D-PSGD} is the dense-consensus reference, enabling nodes to exchange full-precision model information and mix with Metropolis weights over the active topology, following the classical decentralized SGD formulation of~\cite{lian2017can}. Then, {SGP} replaces symmetric mixing with push-sum style communication and therefore serves as the directed/asymmetric-graph counterpart to D-PSGD~\cite{assran2019stochastic}. Furthermore, {CHOCO-SGD} adds communication compression and error-feedback memory, providing a strong compression-aware decentralized baseline without carbon-aware control~\cite{koloskova2019decentralized}. Finally, we include a sparse gossip baseline to represent adaptive peer-to-peer communication with limited fanout and sparsified exchange, following the general communication pattern studied in ~\cite{tang2022gossipfl}. Overall, it should be highlighted that these baselines are complementary in terms of design. D-PSGD tests dense full-precision consensus, SGP tests directed mixing, CHOCO-SGD tests compression with memory-based correction, and Gossip communication tests sparse adaptive peer selection. On the other hand, CARGO extends this design space by introducing carbon-aware control over node activation, edge activation, and compression decisions, while retaining the same underlying decentralized learning substrate.

In greater detail, CARGO uses two fixed runtime presets, held unchanged within each experiment family to avoid per-scenario retuning. The standard preset, used for the client-stress and matched-budget experiments, sets \(d_{\max}=3\), \((\chi_L,\chi_H)=(300,400)\), Top-K \(=0.05\), and \(R=2\). The loss-robust preset, used for packet-loss sensitivity across the three topology regimes, sets \(d_{\max}=2\), \((\chi_L,\chi_H)=(260,340)\), Top-K \(=0.02\), and \(R=3\). Shared controller terms are fixed throughout \(\beta=0\), \(\rho^\star=0.8\), \(S_{\max}=2\), \(\eta_C=\eta_F=0.01\), \(\epsilon=10^{-6}\), and \(\gamma=0.5\), with \(\rho_i(t)\) computed over a fixed 8-round sliding window. This separation is intentional, as the standard participation/budget preset (\(R=2\)) defines a less conservative operating profile, whereas the loss-robust preset (\(R=3\)) uses lower fanout and stronger sparsification. Under harsher packet loss, extra communication and recovery traffic offer lower expected delivered utility per transmitted byte, so the controller is designed to act more conservatively. The presets thus separate a standard operating profile from a loss-robust one rather than tuning CARGO per scenario. The matched-budget study retains the standard preset after budget lock, with fixed delivered-communication and fixed-carbon targets taken from the common feasible overlap across methods and nearest feasible operating points selected under identical seeds and topology/loss traces. 

\subsubsection{Energy and Carbon Accounting}
We use a single analytical accounting model throughout the experiments. Considering a decentralized setting, total cost depends not only on local computation but also on topology-dependent communication, packet loss, compression, and resynchronization, and these effects must be accounted for explicitly. Following standard ML sustainability reporting practice, we compute carbon emissions by combining energy consumption with the corresponding carbon-intensity signal. Specifically, total energy at round $t$ is
\begin{equation}
E_{\mathrm{tot}}(t)=E_{\mathrm{comp}}(t)+E_{\mathrm{comm}}(t),
\end{equation}
and the associated carbon emissions are
\begin{equation}
C_{\mathrm{g}}(t)=\left(\frac{E_{\mathrm{tot}}(t)}{3.6\times 10^6}\right)\mathrm{CI}(t),
\end{equation}
where $\mathrm{CI}(t)$ is expressed in gCO$_2$/kWh. This formulation is consistent with the general energy-to-carbon conversion adopted in ML carbon-accounting tools and prior reporting frameworks~\cite{henderson2020towards,lacoste2019quantifying}. Compute energy is estimated from model workload, device throughput, and active power
\begin{equation}
E_{\mathrm{comp}}(t)=\left(\frac{\mathrm{FLOPs}(t)}{\tau_{\mathrm{dev}}}\right)P_{\mathrm{active}},
\end{equation}
where $\tau_{\mathrm{dev}}$ denotes device throughput (FLOPs/s) and $P_{\mathrm{active}}$ the active power draw (W). Communication energy is estimated from transmitted and received payload bytes using an energy-per-byte link model,
\begin{equation}
E_{\mathrm{comm}}(t)=b_{\mathrm{tx}}(t)\,\epsilon_{\uparrow}\,\kappa_{\ell}+b_{\mathrm{rx}}(t)\,\epsilon_{\downarrow}\,\kappa_{\ell},
\end{equation}
where $b_{\mathrm{tx}}(t)$ and $b_{\mathrm{rx}}(t)$ are transmitted and received bytes, $\epsilon_{\uparrow}$ and $\epsilon_{\downarrow}$ are uplink/downlink energy coefficients, and $\kappa_{\ell}$ is a link-efficiency factor. The same accounting is applied to all methods. This is important in our setting because topology changes may alter communication efficiency and convergence behavior even when the nominal degree budget remains fixed. Since $\mathrm{CI}(t)$ varies across clients and time, carbon reductions need not be proportional to energy reductions.

\subsubsection{Training Protocol}
Clients are locally optimized with Adam, using learning rate $5\times 10^{-4}$, batch size $128$, and no weight decay. Each client performs $1$ local epoch per participation round, and the learning rate follows a cosine schedule. We adopt a fixed-compute evaluation regime in the decentralized setting by constraining the total number of local update steps, typically $84{,}000$, aggregated updates per run) and running periodic evaluation at fixed update intervals every $7{,}000$ updates.

\subsubsection{Evaluation Metrics}
We report three metric groups. Predictive performance is measured through final evaluation $R^2$, root mean square error (RMSE), and MSE loss. Sustainability is measured through total energy (J) and total carbon emissions (gCO$_2$e), each accounting for both compute and communication. Communication efficiency is measured through cumulative communication volume, with attempted and delivered traffic tracked separately under packet loss, together with the empirical effective loss $p_{\mathrm{eff}} = 1 - \frac{\text{delivered}}{\text{attempted}}$. 
All reported values are mean $\pm$ std over five seeds.

\section{Results}
Here, we present the performance comparisons between CARGO and the four baseline schemes. Our evaluation comprises three axes that reflect the deployment objectives of the paper: (i) convergence efficiency, (ii) robustness to client-side instability, and (iii) robustness to unreliable communication.

\subsection{Fidelity and Convergence}
\begin{figure}[!t]
\centering
\includegraphics[width=3.5in]{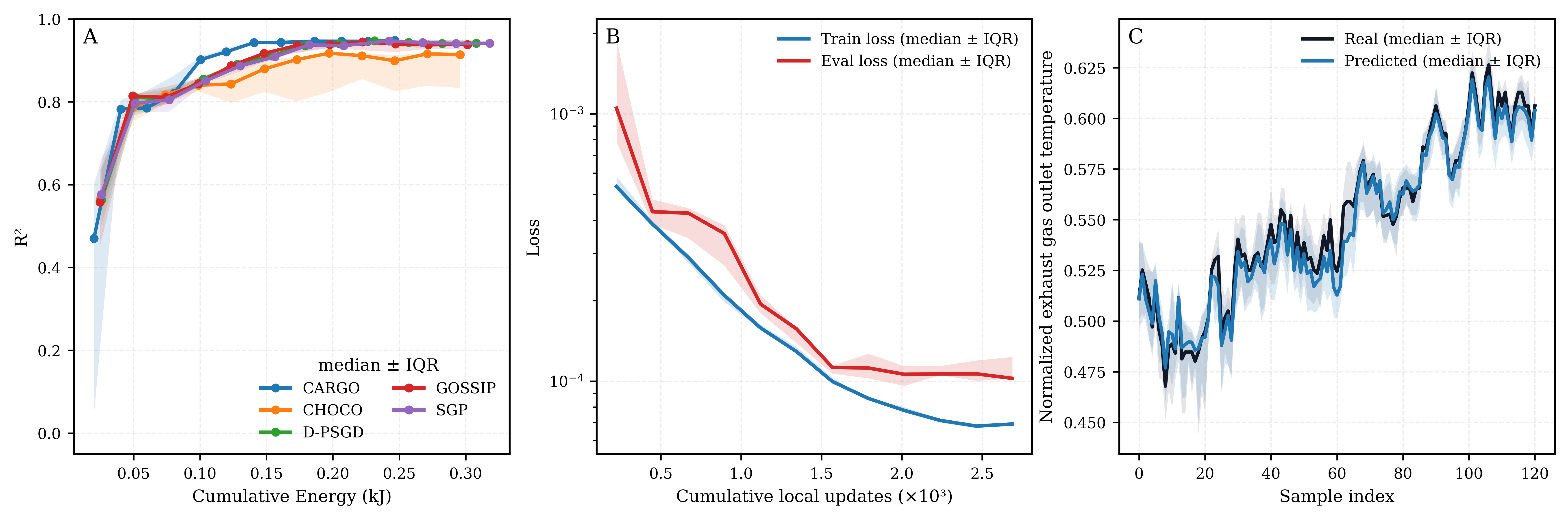}
\caption{Model fidelity and optimization behavior. (A) Predictive performance ($R^2$) versus cumulative compute energy, reported as median with IQR range across seeds/methods. The curve summarizes how quickly each approach reaches the high-$R^2$ plateau per unit energy. (B) Training and evaluation loss trajectories (median $\pm$ IQR) as a function of cumulative local updates, illustrating stable convergence without late-stage divergence. (C) Example time-series segment (median $\pm$ IQR) comparing normalized ground truth and predictions, showing that the learned model tracks temporal dynamics and preserves trend changes under uncertainty.}
    \label{fig:fidelity_convergence}
\end{figure}

Fig.~\ref{fig:fidelity_convergence} characterizes predictive fidelity and optimization dynamics. In Fig.~\ref{fig:fidelity_convergence}A, the $R^2$-energy curves show that the methods reach similar terminal performance levels, but with different energy trajectories. CARGO reaches the high-$R^2$ region earlier along the energy axis, indicating improved energy-to-quality efficiency over the training horizon. The loss trajectories in Fig.~\ref{fig:fidelity_convergence}B decrease smoothly for both training and evaluation, without late-stage instability. Fig.~\ref{fig:fidelity_convergence}C provides a qualitative check on the learned predictor, showcasing that the median prediction follows the median ground-truth trajectory and captures the main trend reversals within the reported dispersion.

\begin{figure}[!t]
\centering
\includegraphics[width=3.5in]{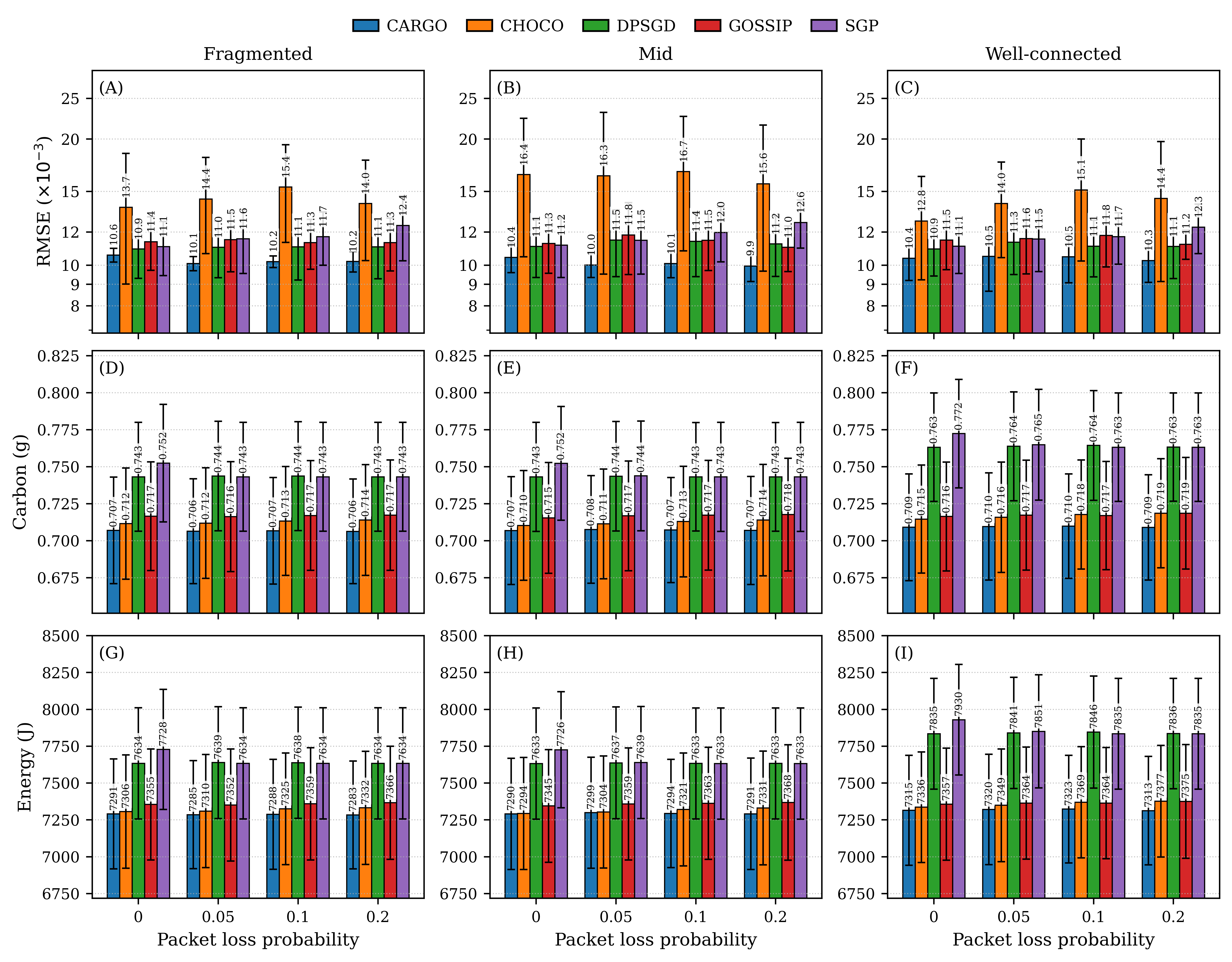}
\caption{Packet-loss robustness across connectivity regimes. Performance and resource profiles under increasing packet-loss probability $p \in \{0,0.05,0.1,0.2\}$ for three topology regimes: fragmented, mid, and well-connected. Top row: final RMSE ($\times 10^{-3}$); middle row: total carbon footprint (g); bottom row: total energy consumption (J). Error bars indicate variability across seeds. The regimes correspond to distinct encounter opportunities derived from the spatiotemporal contact construction (larger communication radius increases connectivity, while coarser temporal binning reduces effective contacts), enabling a controlled evaluation of learning stability and efficiency as network reliability degrades.}
\label{fig:packet_loss_robustness}
\end{figure}

\begin{table}[t]
\centering
\caption{Client-stress grid under mid topology.}
\label{tab:client_stress_mid}
\scriptsize
\setlength{\tabcolsep}{2.0pt}
\renewcommand{\arraystretch}{1.05}
\begin{tabular}{lcccc}
\hline
\textbf{Method} & \textbf{RMSE} & \textbf{Carbon (g)} & \textbf{Energy (J)} & \textbf{Comm. (MB)} \\
\hline

\multicolumn{5}{l}{\textbf{$p_d=0.2$,\quad $f=0.25$}}\\ \hline
CHOCO          & $0.0202\pm0.0076$ & $0.703\pm0.038$ & $7219\pm387$ & $146.9\pm6.7$ \\
D-PSGD         & $0.0094\pm0.0010$ & $0.715\pm0.038$ & $7336\pm387$ & $587.7\pm26.9$ \\
GOSSIP         & $0.0094\pm0.0010$ & $0.725\pm0.039$ & $7448\pm397$ & $846.0\pm8.1$ \\
SGP            & $0.0094\pm0.0010$ & $0.725\pm0.039$ & $7441\pm393$ & $587.7\pm26.9$ \\
\textbf{CARGO} & $0.0095\pm0.0006$ & $0.713\pm0.039$ & $7443\pm403$ & $470.5\pm17.9$ \\

\hline

\multicolumn{5}{l}{\textbf{$p_d=0.2$,\quad $f=0.50$}}\\ \hline
CHOCO          & $0.0143\pm0.0053$ & $0.706\pm0.039$ & $7254\pm399$ & $297.4\pm1.8$ \\
D-PSGD         & $0.0099\pm0.0010$ & $0.726\pm0.039$ & $7455\pm398$ & $1189.6\pm7.3$ \\
GOSSIP         & $0.0098\pm0.0008$ & $0.722\pm0.039$ & $7410\pm404$ & $722.0\pm1.4$ \\
SGP            & $0.0101\pm0.0011$ & $0.736\pm0.039$ & $7557\pm408$ & $1189.6\pm7.3$ \\
\textbf{CARGO} & $0.0094\pm0.0010$ & $0.715\pm0.038$ & $7362\pm399$ & $725.0\pm29.0$ \\
\hline

\multicolumn{5}{l}{\textbf{$p_d=0.2$,\quad $f=1.00$}}\\ \hline
CHOCO          & $0.0145\pm0.0044$ & $0.708\pm0.037$ & $7275\pm383$ & $491.2\pm5.5$ \\
D-PSGD         & $0.0108\pm0.0019$ & $0.737\pm0.037$ & $7573\pm380$ & $1964.7\pm22.1$ \\
GOSSIP         & $0.0109\pm0.0014$ & $0.715\pm0.037$ & $7346\pm382$ & $650.7\pm1.8$ \\
SGP            & $0.0110\pm0.0020$ & $0.745\pm0.037$ & $7655\pm380$ & $1964.7\pm22.1$ \\
\textbf{CARGO} & $0.0101\pm0.0009$ & $0.713\pm0.038$ & $7344\pm390$ & $793.2\pm10.6$ \\

\hline

\multicolumn{5}{l}{\textbf{$p_d=0.5$,\quad $f=0.25$}}\\ \hline
CHOCO          & $0.0290\pm0.0101$ & $0.700\pm0.038$ & $7209\pm392$ & $134.7\pm4.7$ \\
D-PSGD         & $0.0086\pm0.0008$ & $0.715\pm0.038$ & $7356\pm391$ & $538.9\pm18.7$ \\
GOSSIP         & $0.0086\pm0.0008$ & $0.726\pm0.038$ & $7470\pm392$ & $905.5\pm5.2$ \\
SGP            & $0.0086\pm0.0008$ & $0.724\pm0.037$ & $7450\pm381$ & $538.9\pm18.7$ \\
\textbf{CARGO} & $0.0082\pm0.0009$ & $0.709\pm0.039$ & $7307\pm402$ & $367.9\pm19.4$ \\

\hline

\multicolumn{5}{l}{\textbf{$p_d=0.5$,\quad $f=0.50$}}\\ \hline
CHOCO          & $0.0170\pm0.0093$ & $0.705\pm0.038$ & $7229\pm386$ & $235.3\pm3.1$ \\
D-PSGD         & $0.0089\pm0.0008$ & $0.723\pm0.038$ & $7418\pm386$ & $941.3\pm12.5$ \\
GOSSIP         & $0.0087\pm0.0008$ & $0.725\pm0.038$ & $7434\pm384$ & $796.6\pm2.5$ \\
SGP            & $0.0088\pm0.0007$ & $0.732\pm0.041$ & $7510\pm421$ & $941.3\pm12.5$ \\
\textbf{CARGO} & $0.0084\pm0.0005$ & $0.711\pm0.039$ & $7313\pm397$ & $509.6\pm10.0$ \\

\hline

\multicolumn{5}{l}{\textbf{$p_d=0.5$,\quad $f=1.00$}}\\ \hline
CHOCO          & $0.0158\pm0.0057$ & $0.706\pm0.038$ & $7243\pm386$ & $313.3\pm3.0$ \\
D-PSGD         & $0.0105\pm0.0022$ & $0.728\pm0.039$ & $7470\pm393$ & $1253.3\pm11.9$ \\
GOSSIP         & $0.0105\pm0.0016$ & $0.723\pm0.039$ & $7410\pm391$ & $763.6\pm2.6$ \\
SGP            & $0.0106\pm0.0021$ & $0.733\pm0.038$ & $7524\pm384$ & $1253.3\pm11.9$ \\
\textbf{CARGO} & $0.0102\pm0.0005$ & $0.715\pm0.039$ & $7456\pm384$ & $541.1\pm12.5$ \\

\hline
\end{tabular}
\end{table}

\begin{figure}[!t]
\centering
\includegraphics[width=3.55in]{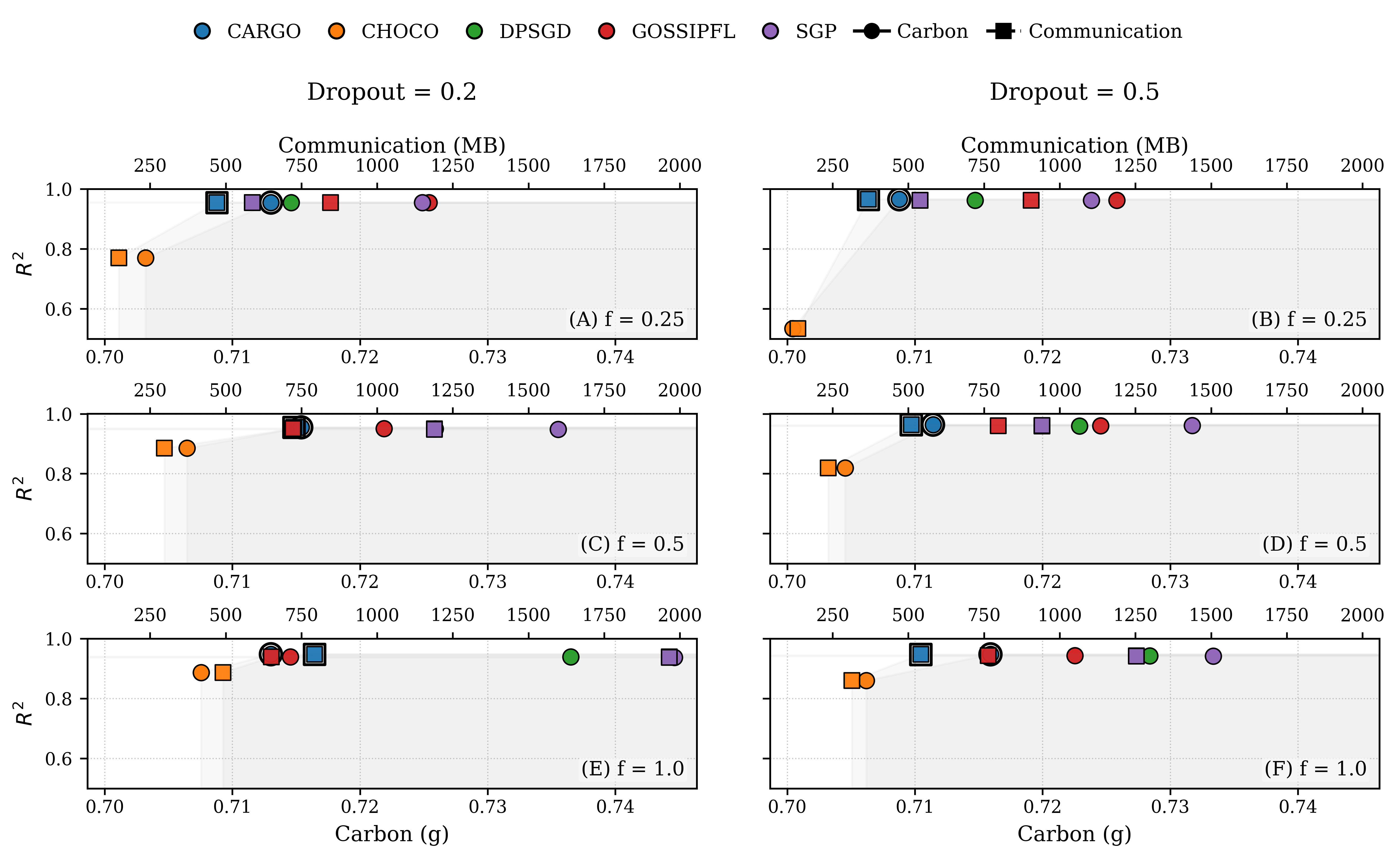}
\caption{Accuracy-efficiency trade-offs under client stress (mid topology). Six panels show $R^2$ versus carbon footprint for availability dropout $p_d \in \{0.2,0.5\}$ and participation fraction $f \in \{0.25,0.5,1.0\}$. Circles use the bottom x-axis (carbon, g), while squares use the top x-axis (communication volume, MB), with both markers plotted at the same $R^2$ for a given method and setting. The figure highlights the practical Pareto structure: CHOCO attains lower carbon/communication at the expense of accuracy, whereas CARGO remains in the high-accuracy region while reducing carbon and communication relative to accuracy-competitive decentralized baselines.}
\label{fig:pareto_r2_carbon_comm}
\end{figure}

\subsection{Packet-Loss Sensitivity Across Connectivity Regimes}
Fig.~\ref{fig:packet_loss_robustness} evaluates sensitivity to increasing packet-loss probability across fragmented, mid, and well-connected regimes. As packet loss increases, RMSE generally rises and variability broadens, especially in the fragmented regime where contact opportunities are limited. Carbon and energy respond through a combination of reduced effective information flow and slower progress within a fixed compute budget. Within this sensitivity analysis, CARGO maintains competitive RMSE while showing favorable carbon and energy profiles relative to the baselines, particularly under higher loss and sparser connectivity where orchestration decisions have a larger effect on effective progress.

\subsection{Client Stress via Dropout and Partial Participation}
Table~\ref{tab:client_stress_mid} quantifies the effect of availability dropout $p_d$ and participation fraction $f$ on error and resource use. Across all settings, CHOCO attains the lowest carbon, energy, and communication volume, but with substantially higher RMSE, consistent with a lower-information operating point. In contrast, the accuracy-competitive decentralized baselines (D-PSGD, sparse gossip, and SGP) achieve low RMSE at higher communication cost. CARGO remains in the high-accuracy regime while reducing resource use relative to these accuracy-competitive baselines. For example, at $(p_d{=}0.5, f{=}0.25)$, CARGO improves RMSE over D-PSGD/SGP (0.0082 vs.\ 0.0086) while reducing carbon (0.709 vs.\ 0.715-0.724) and communication volume (367.9\,MB vs.\ 538.9\,MB). At $(p_d{=}0.2, f{=}0.5)$, it maintains comparable RMSE (0.0094) while reducing carbon relative to D-PSGD/SGP (0.715 vs.\ 0.726-0.736) and substantially reducing communication (725.0\,MB vs.\ 1189.6\,MB), with energy also reduced (7362\,J vs.\ 7455-7557\,J). Fig.~\ref{fig:pareto_r2_carbon_comm} summarizes these operating points and makes the trade-off structure explicit, where CHOCO occupies a low-resource/low-accuracy region, whereas CARGO shifts the high-accuracy region toward lower carbon and communication.

\subsection{Matched-Budget Comparison}
To complement the Pareto-style view, we also evaluate all methods at matched operating budgets in a controlled mid-topology setting. Specifically, all methods are compared at the same feasible carbon budget, $B_{\mathrm{CO2}}=0.6447$~gCO$_2$e, and the same feasible delivered-communication budget, $B_{\mathrm{deliv}}=592.7$~MB. These budgets are defined from the common feasible intersection across methods and seeds, i.e., the largest budget levels for which every method-seed trajectory admits a valid value under the corresponding budget axis. For each method and seed, the performance metric at budget is obtained by linear interpolation along the recorded trajectory. Table~\ref{tab:budget_matched_n5} shows that, at the fixed carbon budget, CARGO attains the strongest mean predictive performance, achieving the highest $R^2$ ($0.941\pm0.012$) and the lowest RMSE ($0.0108\pm0.0011$) among all methods. At the fixed delivered-communication budget, CARGO remains effectively tied with the strongest sparse-gossip baseline and clearly outperforms CHOCO, D-PSGD, and SGP. This comparison is important because it shows that the gains of CARGO are not explained solely by additional flexibility in the controller or by spending more budget, i.e., when carbon or delivered communication is explicitly equalized, CARGO remains best at fixed carbon and top-tier at fixed delivered communication.

\begin{table}[!t] 
\centering 
\caption{Matched-budget comparison at a shared carbon budget and a shared delivered-communication budget.} 
\label{tab:budget_matched_n5} 
\scriptsize
\setlength{\tabcolsep}{2.0pt}
\renewcommand{\arraystretch}{1.05}
\begin{tabular}{lcccc} 
\hline & \multicolumn{2}{c}{\textbf{Fixed CO$_2$}} & \multicolumn{2}{c}{\textbf{Fixed MB}} \\ 
\textbf{Method} & \textbf{$R^2$} & \textbf{RMSE} & \textbf{$R^2$} & \textbf{RMSE} \\ 
\hline 
CHOCO & $0.844\pm0.124$ & $0.0166\pm0.0063$ & $0.849\pm0.116$ & $0.0164\pm0.0060$ \\ 
D-PSGD & $0.933\pm0.026$ & $0.0114\pm0.0021$ & $0.811\pm0.031$ & $0.0193\pm0.0015$ \\ GOSSIP & $0.934\pm0.021$ & $0.0113\pm0.0018$ & ${0.935\pm0.020}$ & ${0.0113\pm0.0017}$ \\ 
SGP & $0.932\pm0.026$ & $0.0115\pm0.0020$ & $0.807\pm0.037$ & $0.0194\pm0.0018$ \\ 
\textbf{CARGO} & ${0.941\pm0.012}$ & ${0.0108\pm0.0011}$ & $0.934\pm0.014$ & $0.0114\pm0.0012$ \\ 
\hline 
\end{tabular} 
\end{table}

\subsection{Complexity Analysis}
Concerning per-round complexity for each model evaluated, 
let $N$ denote the total number of clients, $A\!\le\!N$ the number of active clients in a round (after availability and participation), $d$ the model dimension, $|E_A|$ the number of directed edges in the active-only adjacency, and $|E_S|$ the number of directed edges actually used for communication (method-dependent). All methods share the local training term $O\!\left(A\,C_{\text{local}}(d)\right)$ per round; differences arise from mixing, compression, and scheduling. D-PSGD and SGP perform Metropolis or push-sum mixing over the active topology, yielding graph-aware update cost $O(|E_A|\,d)$ and communication $O(|E_A|\,d)$; depending on the implementation, dense tensor paths may incur an $O(A^2 d)$ compute upper bound while transmitted bytes still scale with the realized $|E_A|$. CHOCO follows the same neighbor scope as the active topology but communicates compressed deltas, giving $O(|E_A|\,d_{\text{eff}} + |E_A|\,C_{\text{comp}}(d,\rho))$ mixing/compression cost and $O(|E_A|\,s(d,\rho))$ communication, where $d_{\text{eff}}=d$ for dense/int8 and $d_{\text{eff}}=\rho d$ for Top-$K$, with $C_{\text{comp}}(d,\rho)$ bounded by $O(d)$ (int8) or $O(d\log(\rho d))$ (Top-$K$) and $s(d,\rho)\in\{O(d),\,O(\rho d)\}$. The GOSSIP baseline selects one neighbor per active node by a local disagreement proxy, so neighbor scoring is $O(|E_A|)$ and communication is $O(|E_S|\,\rho d)$ with $|E_S|\approx A$. CARGO adds a carbon-aware scheduler (score $\rightarrow$ select active $\rightarrow$ pick edges), i.e., node scoring and thresholding is $O(N\log N)$ in the worst case, edge scoring and top-$k$ neighbor selection is $O(|E_A|\log k)$ for out-degree cap $k$, and mixing over the selected subgraph costs $O(|E_S|\,d_{\text{eff}})$ with $|E_S|\approx Ak$; optional resynchronization for waking nodes introduces an amortized $O\!\left(\frac{W}{R}d\right)$ communication term for $W$ resyncing nodes every $R$ rounds.

Overall, the results indicate that CARGO targets the high-accuracy operating region while reducing carbon and communication relative to accuracy-competitive decentralized baselines under both client stress and packet loss. The matched-budget analysis strengthens this interpretation by showing that CARGO remains best at a shared carbon budget and top-tier at a shared delivered-communication budget, rather than benefiting only from a more favorable spending profile. CHOCO remains a lower-resource alternative when accuracy requirements are relaxed, highlighting the expected cost-quality trade-off. Taken together, the evidence supports a Pareto-oriented and budget-aware interpretation of CARGO rather than an “always-best” claim across all metrics and regimes.

\section{Conclusions and Future Directions}
\label{sec:conclusion}

This paper presented CARGO, a carbon-aware orchestration framework for decentralized learning in smart-shipping environments. As maritime learning systems must operate under intermittent connectivity, partial participation, unreliable communication, and increasing pressure to account for resource use and emissions novel collaborative AI frameworks must be provided. In this setting, decentralized training should not be treated as a fixed communication routine, and towards this end, CARGO separates the learning process into a control plane and a data plane, leveraging the former to jointly regulate node participation, edge activation, compression, and recovery behavior. Methodologically, we formulated carbon-aware decentralized maritime PdM as an online control problem over participation and communication decisions. At the systems level, we showed that decentralized learning for smart ships can be improved without changing the local predictor or abandoning established gossip-style updates, with the key being to control when, where, and how communication takes place. Performance evaluation demonstrated that across client-availability stress and packet-loss conditions, CARGO remained in the high-accuracy regime while reducing carbon footprint and communication relative to the accuracy-competitive decentralized baselines. 
The matched-budget analysis further strengthened this point, as when methods were compared at the same feasible carbon budget, CARGO achieved the strongest mean predictive performance, and when compared at the same delivered-communication budget, it remained top-tier and effectively tied with the strongest sparse-gossip baseline. This is important because it shows that the observed advantage is not simply a byproduct of spending more carbon or communication budget. Overall, our work highlights that carbon-aware gossip orchestration is a viable direction for maritime AI, and that resource-aware decentralized learning deserves to be treated as a transportation-systems problem rather than as a minor variant of generic federated learning. 

The present study also has clear scope boundaries. The evaluation is trace-driven rather than a live fleet deployment, focusing on one PdM task and one homogeneous edge-device profile so that differences across methods can be attributed to orchestration rather than to hardware variability. In addition, the experiments prioritize realistic maritime stressors, i.e., changing topology, dropout, and packet loss, over very-large-fleet scaling. Security and privacy mechanisms were also kept outside the present scope, so that the effect of orchestration could be isolated from cryptographic or adversary-robust learning components. 
Future work will therefore extend the framework along three directions: (i) broader fleet-scale evaluation over larger and more heterogeneous maritime deployments, (ii) richer communication models that capture delay and bursty loss in addition to packet drops, and (iii) integration with secure or adversary-robust decentralized learning mechanisms.

\section*{Acknowledgments}
The authors would like to thank Laskaridis Shipping Co.,
Ltd., for data provision.

\bibliographystyle{IEEEtran}
\bibliography{References}

\begin{IEEEbiography}[{\includegraphics[width=1in,height=1.25in,clip,keepaspectratio]{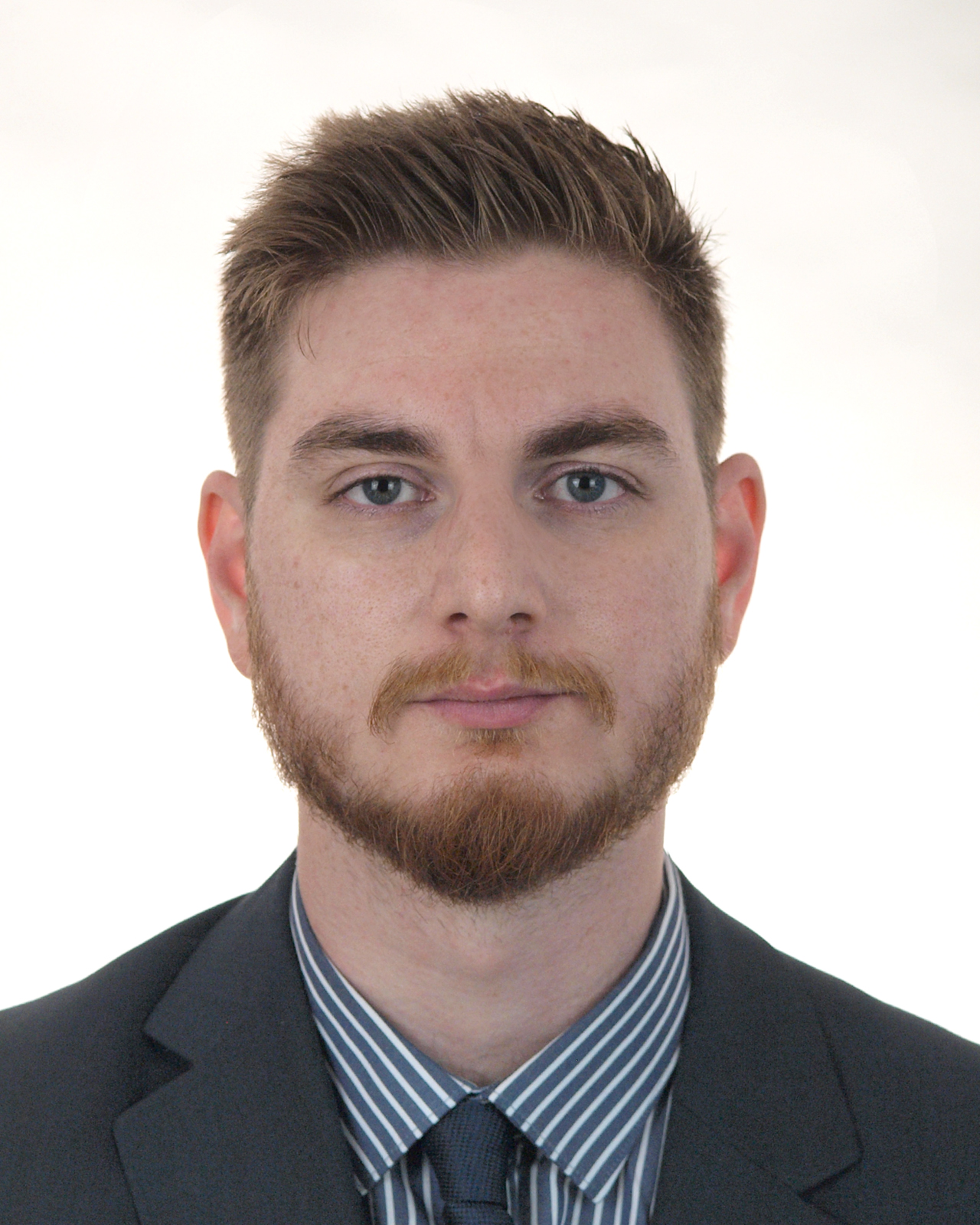}}]{Alexandros S. Kalafatelis} (Graduate Student Member, IEEE) is a Ph.D. candidate at the Dept. of Ports Management and Shipping of the National and Kapodistrian University of Athens (NKUA). He is also a Senior Research Scientist at Four Dot Infinity. He holds a B.Sc. (Hons) in Biomedical Sciences from the University of East London, as well as a B.Eng. in Electrical Engineering and an M.Sc. in Intelligent Management of Renewable Energy Systems, both from NKUA. Part of his research has been conducted in the framework of several European funded R\&D projects, focusing on Predictive Maintenance applications leveraging Federated Learning, with an emphasis on developing secure and optimized aggregation methods tailored to the maritime industry. He serves as a peer reviewer for several IEEE venues, including IEEE TRANSACTIONS ON VEHICULAR TECHNOLOGY and IEEE ACCESS. He also serves as a committee member for the Ship Maintenance, Repair, and Safety Special Interest Group at the Institute of Marine Engineering, Science and Technology (IMarEST) and is the recipient of the Stanley Gray Fellowship. 
\end{IEEEbiography}
\vspace{-4mm} 
\begin{IEEEbiography}[{\includegraphics[width=1in,height=1.25in,clip,keepaspectratio]{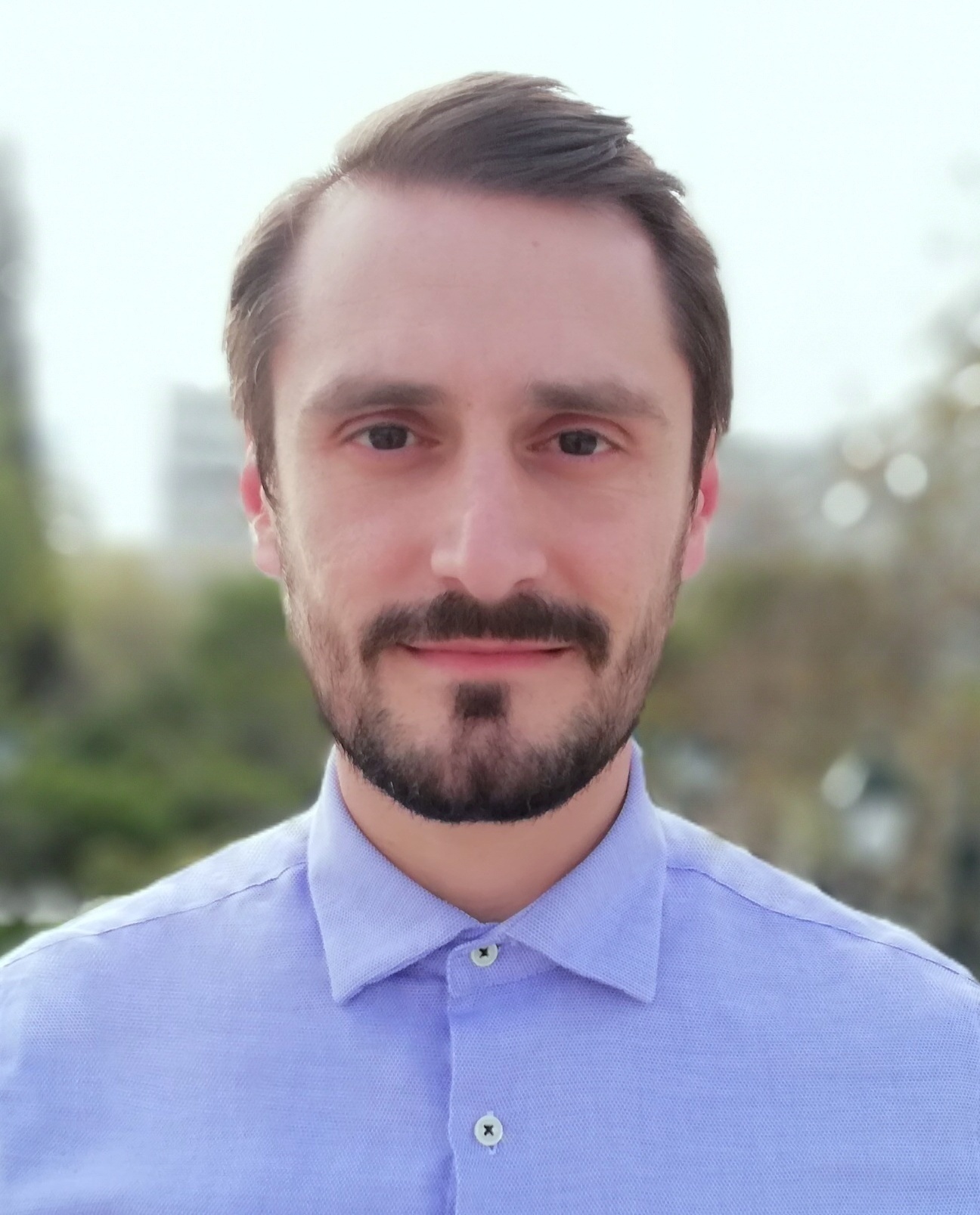}}]{Nikolaos Nomikos} (Senior Member, IEEE) (Senior Member, IEEE)
received the Diploma in electrical engineering and computer technology from the University of Patras, Patras, Greece, in 2009, and the M.Sc. and Ph.D. degrees from the Information and Communication Systems Engineering Department, University of the Aegean, Samos, Greece, in 2011 and 2014, respectively. Since 2025, he has been an Assistant Professor of Mobile and Satellite Communications Systems, Department of Information and Communication Systems Engineering, University of the Aegean, Samos, Greece. Moreover, he is a Project Manager with Four Dot Infinity P.C and a Senior Researcher at the National and Kapodistrian University of Athens. His research interests include cooperative communications, non-orthogonal multiple access, non-terrestrial networks, and machine learning for wireless networks optimization. Prof. Nomikos is an Editor of IEEE TRANSACTIONS ON COMMUNICATIONS and Associate Editor for Frontiers in Communications and Networks. He is a Member of the IEEE Communications Society and the Technical Chamber of Greece. 
\end{IEEEbiography}
\vspace{-4mm} 
\begin{IEEEbiography}[{\includegraphics[width=1in,height=1.25in,clip,keepaspectratio]{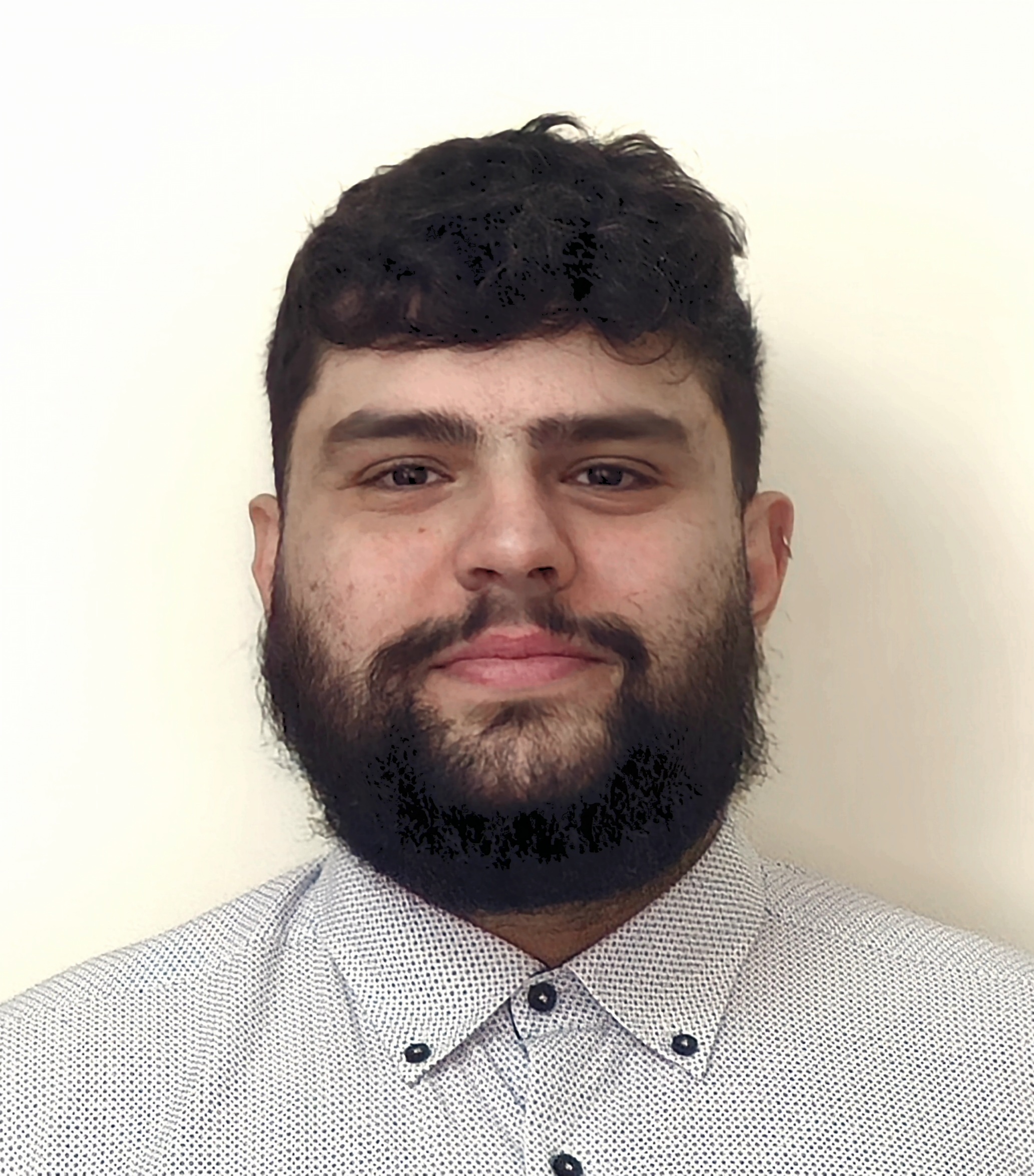}}]{Vasileios Nikolakakis} is a Master's candidate in Intelligent Management of Renewable Energy Systems of the National and Kapodistrian University of Athens (NKUA). He is also a Research Scientist at Four Dot Infinity. He holds a B.Eng in Electrical Engineering from NKUA. His research has been conducted in the framework of several European funded R\&D projects focusing on 5G with primary emphasis on 5G and beyond networks, O-RAN architectures, software-defined networking and secure core network infrastructures. He is also exploring artificial intelligence and machine learning in the scope of network optimization, anomaly detection and resource orchestration, along with DevOps-driven deployment strategies for scalable and reproducible telecom environments.
\end{IEEEbiography}
\vspace{-4mm} 

\begin{IEEEbiography}[{\includegraphics[width=1in,height=1.25in,clip,keepaspectratio]{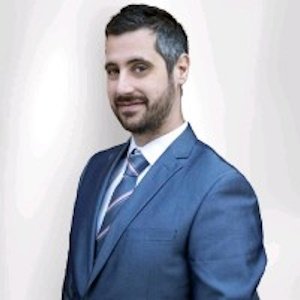}}]{Nikokaos Tsoulakos} graduated from the School of Engineering at the Merchant Marine Academy of Aspropyrgos, attaining the rank of Engineer C’ Class. He then continued obtaining a MEng in Naval Architecture and Marine Engineering and a MSc degree in Marine Science and Technology both from the National Technical University of Athens. Subsequently, he achieved his third BSc degree in Maritime Studies from the University of Piraeus. Currently, he serves as the Innovation \& Technology Manager at Laskaridis Shipping Co. LTD., where he is responsible for overseeing the company’s innovation strategy. 
\end{IEEEbiography}
\vspace{-4mm} 

\begin{IEEEbiography}[{\includegraphics[width=1in,height=1.25in,clip,keepaspectratio]{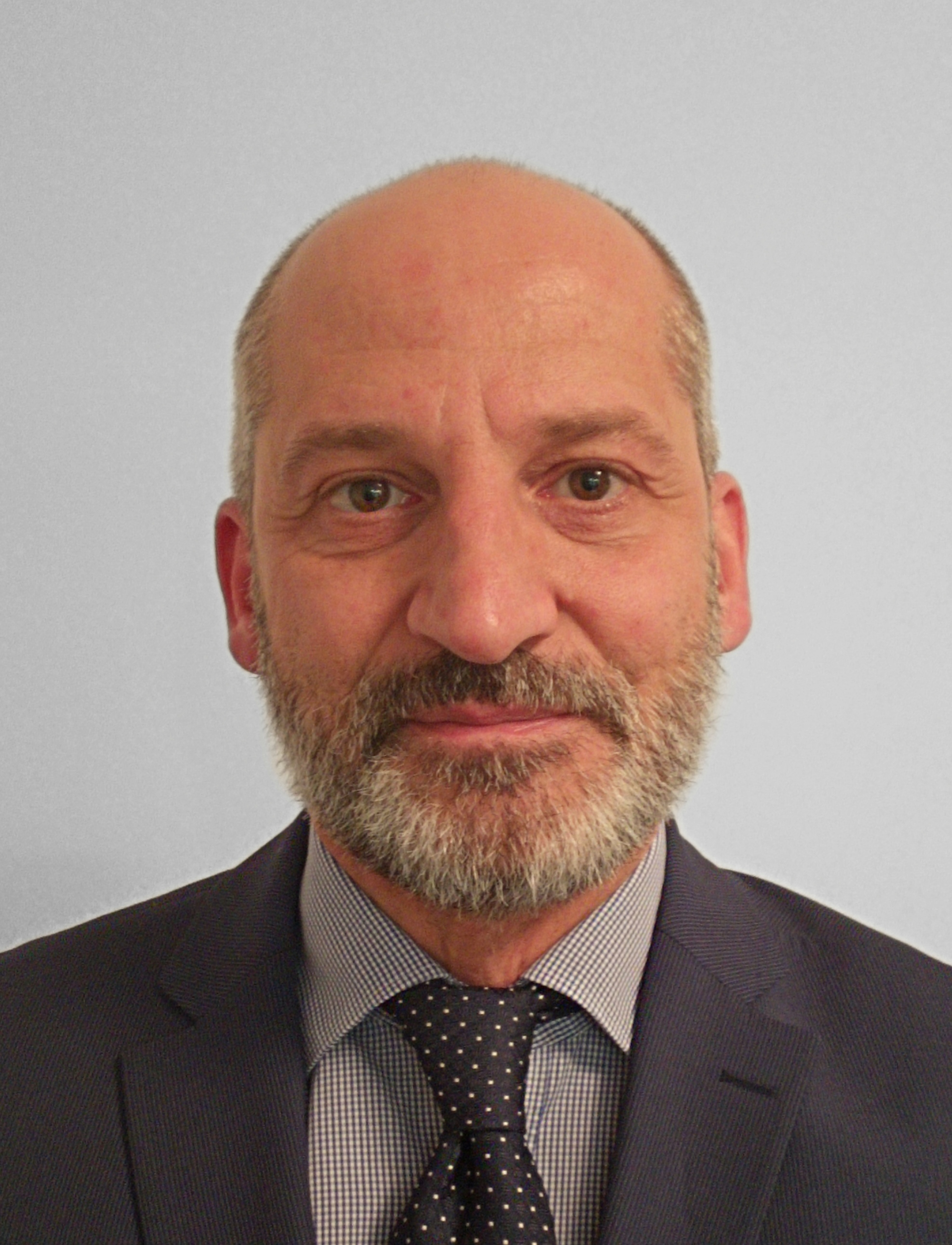}}]{Panagiotis Trakadas} received the Dipl.-
Ing. degree in electrical and computer engineering
and the Ph.D. degree from the National Technical University of Athens (NTUA). In the past,
he was worked at Hellenic Aerospace Industry
(HAI), as a Senior Engineer, on the design of military wireless telecommunications systems, and the Hellenic Authority for Communications Security
and Privacy, where he was holding the position of
the Director of the Division for the Assurance of
Infrastructures and Telecommunications Services Privacy. He is currently a Professor with the National and Kapodistrian University of Athens. He has been actively involved in many EU FP7 and H2020 Research Projects. He has published more than 170 papers in magazines, journals, and
conference proceedings. His research interests include the fields of wireless and mobile communications, wireless sensor networking, network function virtualization, and cloud computing. He is a Reviewer in several journals,
including IEEE TRANSACTIONS ON COMMUNICATIONS and IEEE TRANSACTIONS
ON ELECTROMAGNETIC COMPATIBILITY journals.
\end{IEEEbiography}



\end{document}